\documentclass[runningheads]{llncs}
\usepackage{graphicx}
\usepackage{xcolor}
\usepackage{newfloat}
\usepackage{listings}

\usepackage{bm}
\usepackage{comment}
\usepackage{lipsum}

\usepackage{algorithm}
\usepackage{algpseudocode}

\usepackage{graphicx,float}
\usepackage{adjustbox}
\usepackage{subcaption}  

\usepackage{amsmath,amsfonts,mathtools}
\usepackage{bbold}
\usepackage{cleveref}

\DeclareMathOperator*{\argmin}{arg\,min}

\newtheorem{innercustomlemma}{Lemma}
\newenvironment{customlemma}[1]
  {\renewcommand\theinnercustomlemma{#1}\innercustomlemma}
  {\endinnercustomlemma}
  
\newtheorem{innercustomthm}{Theorem}
\newenvironment{customthm}[1]
  {\renewcommand\theinnercustomthm{#1}\innercustomthm}
  {\endinnercustomthm}

\begin{document}
\title{Scalable  Deep Metric  Learning on Attributed Graphs}
\author{Xiang Li\inst{1} 
\and Gagan Agrawal\inst{2} 
\and Ruoming Jin\inst{3} 
\and Rajiv Ramnath\inst{1} }
\authorrunning{X. Li et al.}
\institute{The Ohio State University, Columbus OH 43210, USA 
\and University of Georgia, Athens GA 30602, USA
\and Kent State University, Kent OH 44242, USA
}
\maketitle              % typeset the header of the contribution

\begin{abstract}

We consider the problem of constructing  embeddings of large attributed graphs and  supporting 
multiple downstream  learning  tasks.  
We develop a graph embedding method, which is based on extending 
deep metric and unbiased contrastive learning techniques to 1) work with attributed graphs,  
2) enabling a mini-batch based approach, and 3)  achieving scalability. 
Based on a multi-class tuplet loss function, we present two algorithms -- DMT for semi-supervised learning and DMAT-i for the  unsupervised case. 
% Our algorithms use Generalized PageRank (GPR) to enrich node features with network 
% topology information, resulting in  scalable construction as well 
% as an informative tuplets for metric distance measurements.  
Analyzing our methods,  we provide a generalization bound for the downstream node classification task and for the first time relate tuplet loss to contrastive learning. 
Through extensive experiments, we show high scalability of representation 
construction, and  in applying the method  for three downstream tasks (node clustering, node classification, and link prediction) 
better consistency over any single existing method.  

\end{abstract}

\section{Introduction} \label{sec:introduction} 
\vspace*{-2.0ex}

Last several years have seen much interest in developing learning techniques on   {\em attributed graphs}, i.e., graphs with  features associated with nodes. 
Such graphs are seen in multiple domains such as recommendation systems~\cite{Ying_2018_GCN_web}, analysis of citation or social networks~\cite{kipf2017semi_GCN,Lazer_2009_life_network},  and  others. 
% A variety of different 
% learning tasks have been considered on such graphs, including  node clustering~\cite{tsitsulin2020graph_dmon,Cui_AGE_2020,ssgc_2021},  node classification~\cite{kipf2017semi_GCN,ssgc_2021,zhu2021_GCA}, and link prediction~\cite{kipf2016variational_VGAE,Cui_AGE_2020}. 
Of particular interest are the deep learning based graph embedding methods~\cite{kipf2016variational_VGAE,ijcai2019_AGC,ssgc_2021,Cui_AGE_2020,zhu2021_GCA} 
that encode graph structural information and node features into low-dimensional representations for multiple downstream tasks. 
% In spite of the progress  from the aforementioned efforts, the quest for representations and methods that are  both general and accurate  (i.e., produce  high quality
% results on different downstream learning tasks)  as well as scalable (capable of handling large graphs) continues. 
Current approaches use Graph Convolutional Networks (GCN)~\cite{zhu2021_GCA} 
or graph filters~\cite{ijcai2019_AGC,ssgc_2021,Cui_AGE_2020}, but either way, 
the methods do not scale to large graphs. 
At a  high level, these graph embeddings are designed with the primary objective 
of pulling  examples with distinct labels apart from each other, while pushing 
the ones sharing the same label closer. It turns out that  the spirit of deep metric learning~\cite{Schroff_2015_FaceNetAU,oh_2016_deep}  is also almost the same, though to date this idea has been primarily applied to learn visual representations~\cite{chen2021_graph_consistency,zhao2021_towards,meyer2018_deep}.
However, besides the challenges of tailoring these  methods for attributed graphs, 
scalability is also a  concern.  
Specifically, 
deep metric learning requires:  1) explicit sampling of tuplets  
such that one or more negative examples is against a single positive example~\cite{oh_2016_deep}, and  2) expensive  
search to increase negative hardness of samples, which is needed 
for enhanced learning power~\cite{Schroff_2015_FaceNetAU,cui_2016_fine,norouzi_2012_hamming}.  

This paper addresses these problems in applying deep metric learning to 
attributed graphs in a scalable fashion. First, we employed an extended version of \textit{multi-class tuplet loss} 
function~\cite{Sohn_N-pair_loss_objective_2016} capable of  working with  multiple positive samples, building on  a similar loss function has been discussed in ~\cite{supervised_CL_2020} for image classification. Next, 
we use (approximate)  Generalized PageRank (GPR) \cite{cwdlydw2020gbp} as a scalable graph  filter,  
which also leads to  a compact node representation and,  as we observe,  increased negative sample 
hardness. Finally, we further achieve scalability 
by mini-batch training; specifically with each batch serving as a 
natural tuplet comprising multiple positive and negative samples; and eliminate the 
cost of sampling. 
%The crux of the idea is to perform pairwise comparison between graph nodes using distance metrics on a manifold such that semantically similar samples are mapped to nearby points while dissimilar ones are projected apart from each other. 
With this basic framework,  we build  multiple algorithms, specifically,  \textbf{D}eep Metric Learning with \textbf{M}ulti-class \textbf{T}uplet Loss (\textbf{DMT})  for semi-supervised learning and  \textbf{DMAT-i} for unsupervised conditions. 

To summarize the novelty of our contributions -- we connect DMAT-i with an extensively applied contrastive loss~\cite{chen_simple_framework_contastive_2020} and theoretically establish  how it  leads to a  bound on the  generalization error of a downstream classification task. 
Equally important, our theoretical analysis explains why contrastive learning is successful for  graph representation learning from a deep metric learning perspective. 
% To summarize the novelty of our contributions --  
% we theoretically proved that the multiclass tuplet loss~\cite{Sohn_N-pair_loss_objective_2016,supervised_CL_2020} is superior to ideal contrastive learning loss (see Lemma 1) and establish a scalable graph embedding method. Our model recognizes multiple positive samples from randomly generated (instead of manually constructed) tuplets and achieves high efficiency and scalability.
On the experimental side, 
we compare  our methods with the state-of-the-art baselines in semi-supervised node classification, node clustering, and link prediction,  and show  more consistent level of accuracy as compared 
to any existing method, and state-of-the-art results in several cases. Finally, 
we also show greater scalability with our methods. 
%as compared to the existing  methods for attributed graphs.  

% Our contribution can be summarized as follows:
% 1) We propose a novel multi-class tuplet loss. With this as the building block, we design several models, DM(A)T and DMAT-i to handle multiple semi-supervised and unsupervised learning tasks.
% 2) We connect DM(A)T with contrastive learning and further investigate the reason behind the success of \textbf{DMAT-i}, which shares the same form of a typical contrastive loss. We theoretically present how \textbf{DMAT-i} leads to bound the generalization error of a downstream classification task. 
% Equally, this explains why contrastive learning can succeed on graph representation learning from metric learning perspective.
% 3) We compare with the state-of-the-art baselines in node clustering, semi-supervised node classification, and link prediction in attributed graphs and achieve a significant performance gain in multiple cases. 
% We demonstrate the great scalability advantages of our method. 

\vspace*{-2ex} 
\section{Preliminaries}\label{sec:preliminary}
\vspace*{-2ex} 

\paragraph{Deep Metric Learning.}
We denote $x \in \mathcal{X}$ as the  input data, with corresponding 
labels  $y \in \mathcal{Y}$. Let $\mathcal{C}$: $\mathcal{X} \to \mathcal{Y}$ be the function of assigning these labels, i.e., $y = \mathcal{C}(x)$.  
In deep metric learning, 
we denote $x^+$ as a {\em positive sample}  of $x$  (i.e., $\mathcal{C}(x^+)=\mathcal{C}(x)$) 
and $x^-$ as  the {\em negative sample} (i.e.,  $\mathcal{C}(x^-) \neq \mathcal{C}(x)$). 
Define $p^+_x(x^\prime)$ to be the probability of observing $x^\prime$ as a positive sample of $x$ and $p^-_x(x^\prime)$ the probability its being a negative sample. 
We assume the class probabilities are uniform such that probability of observing $y$ as a label is $\tau^+$ and probability of observing any different class is $\tau^-=1-\tau^+$. Then the data distribution can be decomposed as $p(x^\prime)=\tau^{+}p^+_x(x^\prime) + \tau^{-}p^-_x(x^\prime)$. 

%\vspace{-8pt}
%\begin{align*}
%        p(x^\prime)=\tau^{+}p^+_x(x^\prime) + \tau^{-}p^-_x(x^\prime)
%\end{align*}

% If the probability of observing $y$ as a label is $\tau$,  then the data distribution can be decomposed as:

% \vspace{-5pt}
% \begin{align*}
%         p(x^\prime)=\tau p^+_x(x^\prime) +  (1 - \tau)  p^-_x(x^\prime)
% \end{align*}

Deep metric learning uses a neural network $f : \mathcal{X} \to \mathbb{R}^{d} $ to learn a $d$-dimensional nonlinear embedding $f(x)$ for each example $x$ based on objectives such as tuplet loss~\cite{Sohn_N-pair_loss_objective_2016} or triplet loss~\cite{Schroff_2015_FaceNetAU}.
\cite{Sohn_N-pair_loss_objective_2016} proposed a \textit{($\mathcal{N}+1$)}-tuplet loss, 
where for a tuplet $(x, x^+,\{x^-_i\}_{i=1}^{N-1} )$ we  optimize  to identify a single positive example from multiple negative examples as: 

\vspace{-7pt}
\begin{equation}
    \label{eq:(N+1)_tuplet}
    \begin{adjustbox}{max width=0.9\columnwidth}
    $
    L^{\mathcal{N}+1}_{\textnormal{tuplet}}(f)=\log \big( 1+\sum_{i=1}^{\mathcal{N}-1}exp \big \{ f(x)^{\top}f(x_i^{-}) - f(x)^{\top}f(x^{+}) \big \} \big) 
    $
    \end{adjustbox}
\end{equation}

This softmax function based objective is {\em  hardness known}  where the hard negative samples receive larger gradients~\cite{Goodfellow-et-al-2016_DL}.

\paragraph{Contrastive Learning.}
In fact, $L^{\mathcal{N}+1}_{\textnormal{tuplet}}$ is mathematically equal to the ideal {\em unbiased contrastive loss} $\widetilde L_{\textnormal{Unbiased}}^{\mathcal{N}+1}(f)$ proposed in~\cite{chuang_2020_debiased}, 
where they introduced: 
\vspace{-5pt}
\begin{equation}\label{eq: unbaised contrast}
\begin{adjustbox}{max width=0.9\columnwidth}
    $
    \widetilde L_{\textnormal{Unbiased}}^{\mathcal{N}+1}(f) = -\log \frac{exp\{{f(x)^{\top} f(x^+)} \}}{exp\{{f(x)^{\top} f(x^+)} \} + (\mathcal{N}-1)\mathbb{E}_{x^- \sim p_x^-} exp\{{f(x)^{\top} f(x^-)} \} } 
    $
\end{adjustbox}
\end{equation}

In contrastive learning, the positive sample (and negative samples) are obtained through perturbation and mainly used in the unsupervised setting (where class label is not available). 
Thus, $p_x^-$ is usually not accessible and  negative samples $x^-_i$ are generated from the (unlabeled) $p(x)$ ~\cite{chuang_2020_debiased}.  Thus, the typical contrastive loss~\cite{chen_simple_framework_contastive_2020} now becomes: 

\vspace{-5pt}
\begin{equation}\label{eq: standard contrast}
\begin{adjustbox}{max width=0.9\columnwidth}
    $
    \widetilde L_{\textnormal{Contrast}}^{\mathcal{N}+1}(f) = -\log \frac{exp\{{f(x)^{\top} f(x^+)} \}}{exp\{{f(x)^{\top} f(x^+)} \} + (\mathcal{N}-1)\mathbb{E}_{x^- \sim p} exp\{{f(x)^{\top} f(x^-)} \} } 
    $
\end{adjustbox}
\end{equation}

Since $x^-_i$ is drawn from $p(x)$, it also has a  probability of $\tau^+$ of being a positive sample. %( Hence $  \widetilde L_{\textnormal{Contrast}}^{N}(f)$ loses accuracy compared with $\widetilde L_{\textnormal{Unbiased}}^{N}$. )
Thus, the contrastive learning is closely related to,  and can even be considered a variant of, deep metric learning, where the positive/negative samples are generated through different perturbation mechanisms. 
To facilitate our discussion, we use the notations $L^{\mathcal{N}+1}_{\textnormal{tuplet}}$ and $\widetilde L_{\textnormal{Unbiased}}^{\mathcal{N}+1}$ interchangeably in the rest of the paper. More related works are reviewed in appendix.

 \vspace*{-2ex} 
\section{Methodology}\label{sec:method}  
\vspace*{-2ex} 
 
\subsection{Problem Statement}
\vspace*{-2ex} 

We are given an attributed graph $\mathcal{G} = (\mathcal{V}, \mathcal{E}, \widetilde{X})$, where $\mathcal{V}=\{v_{1}, v_{2}, \cdots, v_{N}\}$ and $\mathcal{E}$ represent node set and edge set,  respectively, and  $\widetilde{X}$ denotes  the node attributes (i.e., each node is associated with a feature vector).  
Each vertex $v_i$ belongs to a single class (or a cluster) and we apply all notations defined in deep metric learning to graph representations. 
The input data for deep metric learning $\mathcal{X}$ is calculated by a graph filter $\cal{H}$: $\mathcal{X}= {\cal{H}}(\widetilde{X},A)$, where $ A$ is the adjacency matrix. 
Our  objective is to learn an encoder $f : \mathcal{X} \to \mathbb{R}^{d} $ to obtain a $d$-dimensional embedding $f(\mathcal{X})$. 

To develop deep metric learning (or contrastive learning) on graphs, we need to consider and address the  following problems: (1) How to establish a unified approach to cover both semi-supervised and unsupervised settings for graphs? 
(2) How to scale the learning process for large-scale graphs by taking advantage of mini-batch training? 

To elaborate on the second point, the 
existing contrastive learning for graph representation,  
particularly GCA~\cite{zhu2021_GCA}, is built upon a GCN architecture  and uses a typical contrastive loss~\cite{chen_simple_framework_contastive_2020}. It perturbs  the graph topology and node attributes separately, which are fed to GCN to generate augmented views for contrasting. The transformation by GCN  limits both accuracy (due to over-smoothing~\cite{Li_Deeper_into_GCN_2018}) and scalability. 
%To speedup the learning process, we will consider to leverage the linear GCN approach and a ``post-perturbation'' mechanism. 

% We seek an improving scheme especially suitable for graph representation learning in mini-batches: consider each of shuffled node batches as a natural tuplet, in which we aim to identify multiple positive examples simultaneously from all their negative counterparts. 1) integrate graph information into high-quality representations to enable mini-batch training; 2) use each batch as a natural tuplet, on which deep metric learning is conducted by identifying multiple positive examples simultaneously from all their negative counterparts. Motivated by this idea, we propose a novel loss function $L_{DMT}$ and an efficient deep metric learning framework where no explicit tuplet construction is needed. We illustrate the accuracy of $L_{DMT}$ in capturing semantics by theoretically showing $L_{DMT} \leq L^{\mathcal{N}+1}_{\textnormal{tuplet}}(f)$ given same size of tuple.

% $\widetilde L_{\textnormal{Unbiased}}^{N}(f) $ is considered as an ideal contrastive loss~\cite{chuang_2020_debiased}. 

\vspace*{-2pt} 
\subsection{DMT Algorithm} 
\vspace*{-2ex} 

We first propose the learning framework, \textbf{D}eep Metric Learning with \textbf{M}ulti-class \textbf{T}uplet (\textbf{DMT}), for semi-supervised node classification task.  
By applying a multiclass tuplet loss~\cite{Sohn_N-pair_loss_objective_2016,supervised_CL_2020} which can recognize multiple positive samples from the tuplet, DMT addresses the aforementioned batch and scalability problem with the following distinguishing advantages: 1) high scalability and efficiency is achieved by using each shuffled node batch as a natural tuplet -- this choice also  alleviates the  need for explicit (and expensive) sampling; 2) enhanced and faster representations construction through graph filtering, which we show later increases {\em negative sample hardness}. 

Specifically,  \textbf{DMT} employs a GPR-based graph smoothing filter $\mathcal{H}$ -- 
as described earlier, the goal is to smooth node attributes $\widetilde{X}$ by graph structure via $\mathcal{X}= {\cal{H}}(\widetilde{X},A)$ such that each $x \in \mathcal{X}$ contains information from its neighborhood as well. The details of this filtering, and how it can be done on 
large graphs, is presented  in the appendix.
This approach can also help increase negative sample hardness, 
a property that has been shown to  accelerate training and enhance the discriminative power~\cite{Schroff_2015_FaceNetAU,cui_2016_fine,norouzi_2012_hamming} - 
details  again are captured in the appendix. 

% However, following spirit of semi-supervised learning~\cite{kipf2017semi_GCN}, this enables us to utilize limited train nodes (with labels) to learn an embedding and infer node classification for the entire graph. 
% We also show (in appendix) that 
% compared with $\widetilde{X}$, our input $\mathcal{X}$ also increases negative sample hardness, 
% a property that has been shown to  accelerate training and enhance the discriminative power~\cite{Schroff_2015_FaceNetAU,cui_2016_fine,norouzi_2012_hamming}.  

DMT employs an extended version of the multi-class tuplet loss from the deep metric learning~\cite{Sohn_N-pair_loss_objective_2016}.
Training is conducted in mini-batches and we consider each train batch $\mathcal{X}_{B}$ of size $\cal{B}$ as a ${\cal{B}}$-tuplet $ (x, \{x^+_i\}_{i=1}^{m}, \{x^-_i\}_{i=1}^{q} )$
with $m$ positive samples $x^{+}$ and $q$ negative samples $ x^{-}$ of $x$ respectively 
($m$ and $q$ are batch dependent). 
Furthermore, we define $h(x, x^\prime;f)  = exp\{\frac{f(x)^{T} \cdot f(x^\prime)}{t}\}$, where we apply the cosine similarity as a metric distance such that each feature vector $f(x)$ is normalized before performing the Cartesian product.
Temperature $t$ is the radius of hypersphere where the representations lie~\cite{wang_2020_hypersphere} and can control penalty degree on hard negative samples as inspired by~\cite{wang_2021_CL_behavior}.   

Now, the \textit{multi-class tuplet loss} function is: 

\vspace{-5pt}
\begin{equation}
    \label{eq:DMT_pairwise_obj}
    \begin{adjustbox}{max width=0.9\columnwidth}
    $
        L^{\textnormal{m,q}}_{\substack{\textnormal{DMT}}}(x;f) = 
        -\log \frac{ h(x,x;f) + \sum_{i=1}^m  h(x, x_i^+;f)   
        }
        {
        h(x,x;f) + \sum_{i=1}^{m} h(x, x_i^+;f) +
        \sum_{i=1}^{q} h(x, x_i^-;f)
        }  
    $
    \end{adjustbox}
\end{equation}

Here,  $x$ is counted as one positive sample of itself to avoid zero-value inside the \textit{log} function. 
The loss function above shares a close mathematical form of supervised contrastive loss as proposed in ~\cite{supervised_CL_2020} and enables us to create  efficient  mini-batch versions, 
while preserving the essential ideas behind  metric  or contrastive learning. 
One important aspect is because the function  can work with varying $m$ and $q$ across batches, we can simply use all the  positive and negative samples associated with any given batch. 

\vspace*{-2ex} 
\subsection{DMAT-i Algorithm}
\vspace*{-2ex} 

In the unsupervised cases, $\{x^+_i\}$ and $\{x^-_i\}$ are no longer recognizable. 
To deal with this problem, we adopt the idea of contrastive learning, which includes  multiple views of graph embeddings through {\em augmentation}, while assuming that the labeling still exists initially (thus, drawing  from the  deep metric learning framework). Then, we will show we can drop out the labels of the loss, which leads to the format of the contrastive learning loss. 

Specifically, for one batch of samples $\mathcal{X}_B$ of size ${\cal{B}}$ together with their augmented counterparts, we have a $2{\cal{B}}$-tuplet  $(x, \bar{x}, \{x^+_i\}_{i=1}^m,  \{x^-_i\}_{i=1}^q )$ with $m$ positive pairs and $q$ negative pairs  -- here,  $\bar{x}$ denotes the augmented counterpart (trivial positive sample) of $x$. Thus, we introduce an immediate DMAT tuplet loss $L_{\substack{\textnormal{DMAT}}}^{\textnormal{m,q}}(x, \bar{x};f)$ following the similar form of Eq.\ref{eq:DMT_pairwise_obj}:

\vspace{-5pt}
\begin{equation}\label{eq:DMAT_loss}
    \begin{adjustbox}{max width=0.9\columnwidth}
    $
    L_{\substack{\textnormal{DMAT}}}^{\textnormal{m,q}}(x, \bar{x};f)=-\log \frac{ h(x,\bar{x};f) + \sum_{i=1}^m  h(x, x_i^+;f)}{h(x,\bar{x};f) + \sum_{i=1}^m  h(x, x_i^+;f) + \sum_{i=1}^q  h(x, x_i^-;f)  }
    $
    \end{adjustbox}
\end{equation}

% Next, we further incorporate the ideas of contrastive learning to include  multiple views of graph embeddings through {\em augmentation}, 
% resulting in  additional  positive samples,

% Node classification requires the model to tolerate intra-class variations among each class.  
% To this end, we further extend DMT into a new model \textbf{DMAT}.  The idea is to 
% include  multiple views of graph embeddings through {\em augmentation}, 
% resulting in  additional  positive samples, in turn leading to  a more robust embedding.  

\begin{figure}
     \centering
     \includegraphics[width=0.8\linewidth]{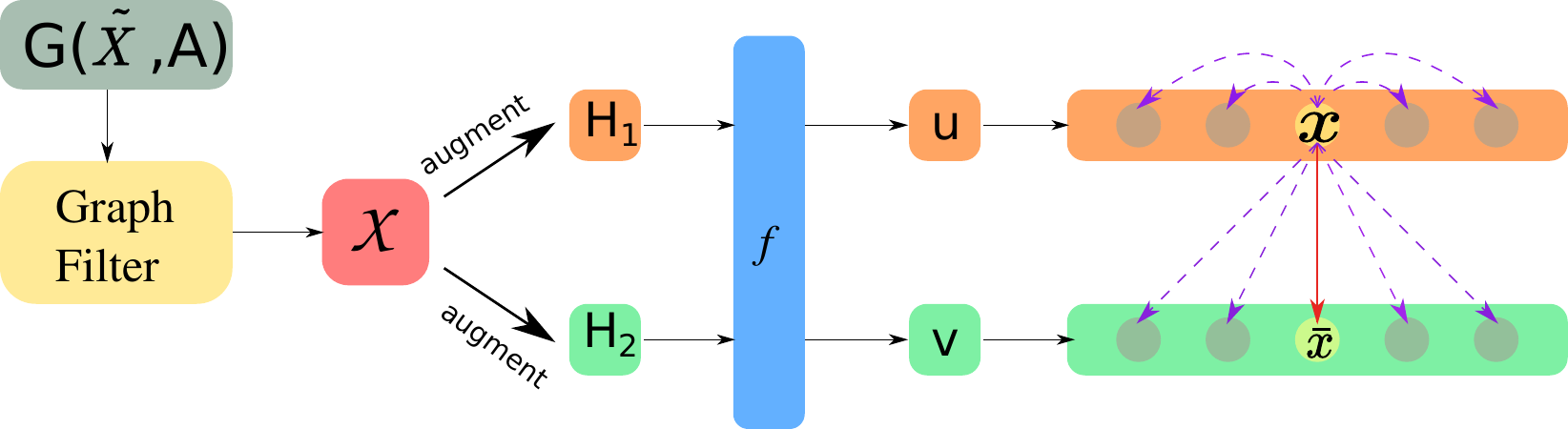}
     \caption{Schematic of DMAT-i architecture. The graph filter generates smoothed node attributes $\mathcal{X}$ by incorporating graph structural information. 
    A pair of views ($H_{1}, H_{2}$) of $\mathcal{X}$ are produced by augmentation and fed to the subsequent encoder $f$ to generate latent representations $U=f(H_{1})$ and $V=f(H_{2})$. Metric distance measurement is performed on $U \bigcup V$. For each sample $x \in U$, its counterpart $\bar{x} \in V$ is the only recognizable positive sample.
     }
     \label{fig: contrastive training arch}
\end{figure}
\vspace*{-4ex} 

%%% ================== DMAT-i =================
%\paragraph{ DMAT-i (Unsupervised Learning)}
Next, we extend DMAT to unsupervised cases where $\{x^+_i\}$ and $\{x^-_i\}$ are no longer recognizable. Here, the resulting method, DMAT-i, involves further simplification by extracting $\bar{x}$ as the only positive sample of $x$ while ignoring all other positive ones. The loss function 
 is (mathematically equal to Eq.\ref{eq: standard contrast}): 

\begin{equation}\label{eq:DMAT-i_loss}
    \begin{adjustbox}{max width=0.85\columnwidth}
    $
    L^{\textnormal{m,q}}_{\substack{\textnormal{DMAT-i}}}(x,\bar{x};f)=-\log \frac{ h(x,\bar{x};f)}{h(x,\bar{x};f) + \sum_{i=1}^m  h(x, x_i^+) + \sum_{i=1}^q  h(x, x_i^-)  }
    $
    \end{adjustbox}
\end{equation}

Note ${\{ x^+_i \}}_{i=1}^m$ and ${\{ x^-_i \}}_{i=1}^q$ are explicitly denoted for ease of analysis, 
but they remain unknown during the  training.   Eq.~\ref{eq:DMAT-i_loss} is in fact calculated without knowing any labels as:

\begin{equation}
    \begin{adjustbox}{max width=0.95\columnwidth}
    $
 L^{\textnormal{m,q}}_{\substack{\textnormal{DMAT-i}}}(x,\bar{x};f) = \log \big \{ {\sum_{\substack{x^{\prime} \in \mathcal{X}_{B} \\ x^{\prime} \neq x} }  h(x, x^{\prime};f)}/{h(x,\bar{x};f)} \big \} \nonumber
    $
    \end{adjustbox}
\end{equation}  

\noindent \textbf{Complete Algorithm:}
The general idea is illustrated in Figure~\ref{fig: contrastive training arch}. Augmented views are generated on the fly from $\mathcal{X}$ by masking certain columns -- the 
consequence is that the  node features and structural information (encoded inside $\mathcal{X}$) are ``distorted'' 
simultaneously. A subsequent DNN based module can abstract information and perform metric similarity measurements (as in  Eq. \ref{eq:DMAT-i_loss}) between each pair of views. 
In real implementation, we use $\mathcal{X}$ as the anchor view and each augmented view as the counterpart to calculate an average of training loss. Thus, the encoder will be optimized to learn robust characteristics of representations across different views. 
The overall objective to be maximized is defined as the average  agreement $L_{\substack{\textnormal{DMAT-i}}}(x, \bar{x};f)$ over all interchangeable view pairs as follows: 

\vspace{-2pt} 
\begin{equation}\label{eq:DMAT-i_loss_multi_view}
    \begin{split}
        \mathbf{J} = \frac{1}{2{\cal{B}}}\sum_{x \in {{\mathcal{X}}_B}}[L^{\textnormal{m,q}}_{\substack{\textnormal{DMAT-i}}}(x, \bar{x};f) + L^{\textnormal{m,q}}_{\substack{\textnormal{DMAT-i}}}(\bar{x},x;f)]
    \end{split}
\end{equation}   
\vspace{-2pt} 
 
The entire training process is presented in Algorithm~\ref{alg:DMAT Train}. As input, $\mathcal{X}$
is generated using random-walk based GnnBP (Graph neural network via Bidirectional Propagation~\cite{cwdlydw2020gbp}) as graph filtering. 
In line 3, multiple ($n_{view}$) augmented embedding will be generated from one batch of filtered feature $\mathcal{X}_{B}$ by masking certain columns in $\mathcal{X}_{B}$.
In line 5, the  generated graph embedding views will be input into the DNN based encoder $f$ to produce the latent representations. The deep metric learning in line 6 is  performed in batches between encoded representations $u$ of the anchor view $\mathcal{X}_{B}$ and $v$ of each augmented view ${H}_{B}$. The obtained embedding $Z$ in line 8 will be used for the  downstream learning tasks.

\begin{algorithm}
    \begin{minipage}{0.9\linewidth}
	\caption{DMAT-i Training} 
	\label{alg:DMAT Train}
	\begin{itemize}
    	\Statex{\textbf{Input data}: GnnBP filtered attributes $\mathcal{X}$, Graph $G$, number of views: $n_{view}$}
	\end{itemize}
	\begin{algorithmic}[1]
    \For{$epoch = 1, 2, \cdots $} 
	    \For{$\mathcal{X}_{B}$ in $\mathcal{X}$ }
	        \State{Generate $n_{view}$ augmented views of $\mathcal{X}_{B}$: $\{{H}_{B}\}$  }
	        \For{$i = 1, 2, \cdots $ }
	            \State{$ u \gets f(\mathcal{X}_{B})$; $ v \gets f({H}_{B}^{i})$ }
	            \State{Compute multi-class tuplet loss $\mathbf{J}$ (Eq.\ref{eq:DMAT-i_loss_multi_view})}
	        \EndFor
	        \State{SGD update on $f$ to  minimize $\mathbf{J}$}
	    \EndFor
	\EndFor
    	\State{$Z \gets f(\mathcal{X})$}
	\end{algorithmic}
    \end{minipage}
\end{algorithm}

\vspace{-2pt}

\vspace*{-2ex} 
\section{Theoretical Analysis}
\label{sec:theory}
\vspace*{-2ex} 

\paragraph{DM(A)T and Contrastive Learning}

$\widetilde L_{\textnormal{Unbiased}}^{\mathcal{N}+1}(f)$ (Eq.~\ref{eq: unbaised contrast}) contrasts one positive sample against multiple negative samples and has been  recognized as the  ideal loss to optimize~\cite{chuang_2020_debiased}. $L_{\substack{\textnormal{DM(A)T}}}^{\textnormal{m,q}}(f)$ improves $\widetilde L_{\textnormal{Unbiased}}^{\mathcal{N}+1}(f)$ by recognizing multiple positive samples at the same time. It turns out that it  can be shown as a lower bound of $ L_{\textnormal{Unbiased}}^{\mathcal{N}+1}(f)$, specifically:

\begin{customlemma}{1}\label{lemma: DMAT inequality}

For any embedding $ f $, given the same size of tuplets sharing one positive sample ${x_{0}^{+}}$, i.e. $(x, x_{0}^+,\{x^-_i\}_{i=1}^{N-1})$ for $L_{\textnormal{Unbiased}}^{\mathcal{N}+1} $ and $(x, x_{0}^{+}, \{x^+_i\}_{i=1}^m,  \{x^-_i\}_{i=1}^q )$ for $   L_{\textnormal{DM(A)T}}^{\textnormal{m,q}} $, we have:
$   L_{\textnormal{DM(A)T}}^{\textnormal{m,q}}(f) 
      \leq  \widetilde L_{\textnormal{Unbiased}}^{\mathcal{N}+1}(f) $
\end{customlemma}

Now, as we know,  both $\widetilde L_{\textnormal{Unbiased}}^{\mathcal{N}+1}(f)$ and $L_{\substack{\textnormal{DM(A)T}}}^{\textnormal{m,q}}(f)$ require  $p_x^+$ and $p_x^-$,  which can only be accessed from training data (i.e., during supervised learning). 
For unsupervised conditions, our $L_{\textnormal{DMAT-i}}^{\textnormal{m,q}}$ considers $\bar{x}$ as the only available positive sample. Next, we will show how $L_{\textnormal{DMAT-i}}^{\textnormal{m,q}}$ contributes to a downstream learning task.

\paragraph{DMAT-i Generalization Bound on Node Classification}

We relate $L_{\textnormal{DMAT-i}}^{\textnormal{m,q}}$ to a supervised loss and present how $L_\textnormal{DMAT-i} ^{\textnormal{m,q}}$ leads to a generalization bound for a supervised node classification task.
Consider a supervised node classification task with $K$ classes, we fix the embedding $f(\mathcal{X})$ from DMAT-i representation learning and train a linear classifier $\psi(\mathcal{X}) = f(\mathcal{X})W^\top $ with the standard multi-class softmax cross entropy loss $L_{\textnormal{Softmax}}(\psi)$. We define the supervised loss for the representation $f(\mathcal{X})$ as:
$ L_{\textnormal{Sup}}(f) = \inf_{W \in \mathbb{R}^{K \times d}} L_{\textnormal{Softmax}}(fW^\top)$ 
%\end{align*}

%\begin{align*}
%    L_{\textnormal{Sup}}(f) = \inf_{W \in \mathbb{R}^{K \times d}} %L_{\textnormal{Softmax}}(Wf).
%\end{align*}

\cite{chuang_2020_debiased} has proved $\widetilde L_{\textnormal{Unbiased}}^{\mathcal{N}+1}(f) $ as an upper bound of $L_{\textnormal{Sup}}(f)$. What we contribute here is 
to bound the difference between $\widetilde L_{\textnormal{Unbiased}}^{\mathcal{N}+1}(f) $ and $L_{\textnormal{DMAT-i}}^{\textnormal{m,q}}$.

\begin{customthm}{1}\label{thm: bound objective diff}
For any embedding $f$ and same size of tuplets, 
\vspace{-3pt}
\begin{equation}
\begin{adjustbox}{max width=\columnwidth}
$
\left | \widetilde L_{\textnormal{Unbiased}}^{\mathcal{N}+1}(f) - L_{\substack{\textnormal{DMAT-i}}}^{\textnormal{m,q}}(f) \right |
    \leq \sqrt{ \frac{2(e^3-e)(\tau^{0})^{2} \pi}{m}} +   \sqrt{ \frac{2(e^3-e)(\tau^{-})^{2} \pi}{q}} \nonumber 
$
\end{adjustbox}
\end{equation}

\vspace{-5pt}

\begin{equation}\label{eq: tau0}
\begin{adjustbox}{max width=\columnwidth}
$
\tau^{0} = \tau^{+}  \bigg ( \frac{\big | \frac{1}{m} \sum_{i=1}^m  h(x, x^+_i)  - \mathbb{E}_{\substack{x^- \sim p_x^-} } h(x,x^-) \big |}{\big | \frac{1}{m} \sum_{i=1}^m  h(x, x^+_i)  - \mathbb{E}_{\substack{x^+ \sim p_x^+} } h(x,x^+) \big |} \bigg )
$
\end{adjustbox}
\end{equation}

\end{customthm}

where $\sum_{i=1}^m  h(x, x^+_i)$ represents the positive samples unrecognized by $L_{\textnormal{DMAT-i}}^{\textnormal{m,q}}$, i.e., \textit{false negative samples}. Hence $\tau^{0}$ covers the side effects from these false negatives and an empirical evaluation in appendix has shown reasonable small values of $\tau^{0}$ for most samples across our experimental datasets.

In practice, we use an empirical estimate $\widehat L_{\substack{\textnormal{DMAT-i}}}^{\textnormal{m,q}}(f)$ over $N$ data samples $x \in \mathcal{X}$, each sample with a tuplet $(x, \bar{x}, \{x^+_i\}_{i=1}^m,  \{x^-_i\}_{i=1}^q )$. The optimization process learns an empirical risk minimizer $\widehat{f} \in \argmin_{f \in \mathbb{F}} 
L_{\textnormal{DMAT-i}}^{\textnormal{m,q}}(f)$ 
from a function class $\mathbb{F}$. 
The generalization depends on the \emph{empirical Rademacher complexity} 
$\mathcal{R}_\mathcal{S}(\mathbb{F})$ of $\mathbb{F}$ with respect to our data sample $\mathcal{S} = \{ x_j, \bar{x}_j, \{ x^{+}_{i,j}\}_{i=1}^m, \{ x^{-}_{i,j}\}_{i=1}^q  \}_{j=1}^N$. Let $f_{|\mathcal{S}} = (f_k(x_j), f_k(\bar{x}_j), \{f_k(x^{+}_{i,j}) \}_{i=1}^m, \{f_k(x^{-}_{i,j}) \}_{i=1}^q)_{j \in [N], k \in [d]} \in \mathbb{R}^{(m+q+2)dN}$ be the restriction of $f$ onto $\mathcal{S}$, using $[N] = \{1,\ldots, N\}$ and $[d] = \{1,\ldots, d\}$. Then $\mathcal{R}_\mathcal{S}(\mathbb{F})$ is defined as:
$  \mathcal{R}_\mathcal{S}(\mathbb{F}) :=  \mathbb{E}_\sigma \sup_{f \in \mathbb{F}} \langle \sigma, f_{|\mathcal{S}}\rangle $ 
%\end{align*}
%\begin{align*}
% \mathcal{R}_\mathcal{S}(\mathbb{F}) :=  \mathbb{E}_\sigma \sup_{f \in \mathbb{F}} \langle %\sigma, f_{|\mathcal{S}}\rangle
%\end{align*}
where $\sigma \sim \{ \pm 1\}^{(m+q+1)dN}$ are \textit{Rademacher random variables}. We provide a data dependent bound from $L_{\substack{\textnormal{DMAT-i}}}^{\textnormal{m,q}}(f)$ on the downstream supervised generalization error as follows.

\vspace{-2pt} 
\begin{customthm}{2}\label{thm: bound sup diff}

With probability at least $1-\delta$, for all $f \in \mathbb{F}$ and $q \geq K-1$,
\vspace{-3pt} 
 \begin{equation}
\begin{adjustbox}{max width=\columnwidth}
$
L_{\textnormal{Sup}}(\hat{f})  \leq  L_{\substack{\textnormal{DMAT-i}}}^{\textnormal{m,q}}(f) + \mathcal{O} \left (\tau^{0} \sqrt{ \frac{1}{m}} + \tau^{-} \sqrt{\frac{1}{q}} + 
      \frac{\lambda \mathcal{R}_\mathcal{S}(\mathbb{F})}{N}+  \Gamma\sqrt{\frac{\log{\frac{1}{\delta}}}{N}} \right ) \nonumber 
$
\end{adjustbox}
\end{equation}
 
where $\lambda = \frac{(m+q)e}{m + q + e}$ 
and $\Gamma = \log (m+q) $.

\end{customthm}
\vspace{-2pt} 

The bound states that if the function class $\mathbb{F}$ is sufficiently rich to contain  embeddings  for which $L_{\substack{\textnormal{DMAT-i}}}^{\textnormal{m,q}}$ is small, then the representation encoder $\hat f$, learned from a large enough dataset, will perform well on the downstream classification task.
The bound highlights the effects caused by the false negative pairs with the first term and also highlights the role of the inherent positive and negative sample sizes $m$ and $q$ per mini-batch in the objective function. The last term in the bound grows slowly with $m+q = 2\mathcal{B}-2$, but the effect of this on the generalization error is small if the dataset size $N$ is much larger than the batch size $\mathcal{B}$, as is common. 
%the case. 

\vspace{-4pt} 
\section{Experimental Results}  
\label{sec:expr}

% \begin{footnotesize} 
% \begin{table}[tbh]
%     % \centering
%     \caption{Datasets Statistics}
%     \begin{adjustbox}{max width=\columnwidth,center}
%     \begin{tabular}{lrrrr}
%     \hline
%     dataset  & Nodes & Classes & Features & Edges \\ 
%     \hline
%     ACM       & $ 3025  $  & $ 3 $  & $ 1870 $   & $13,128$ \\
%     DBLP      & $ 4058 $   & $ 4 $  & $ 334 $    & $3528$ \\
%     Citeseer  & $ 3327  $  & $ 6 $  & $ 3703 $   & $4732$ \\
%     Cora      & $ 2708  $  & $ 7 $  & $ 1433 $   & $5429$ \\
%     Pubmed    & $ 19717  $  & $ 3 $  & $ 500 $   & $44,338$ \\
%     Amazon Photo  & $ 7650 $   & $ 8 $ & $ 745 $    &  $71,831$ \\
%     Coauthor CS     & $ 18333 $  & $ 15 $ & $ 6805 $  &  $81,894$   \\
%     Coauthor PHY     & $ 34493 $  & $ 5 $ & $ 8415 $  &  $247,962$  \\
%     % Ogb-arxiv  & $ 169343 $  & $ 40 $ & $ 128 $  &  $ 1,166,243	 $  \\
%     % Reddit   & $ 232965 $  & $ 41 $ & $ 602 $  &  $114,615,892$  \\
%     % Friendster   &  $6.5 \times  10^7$   & $ 7 $ & $ 40 $  &  $1.8 \times 10^9$  \\
%     \hline
%     \end{tabular}
%     \end{adjustbox}
%     \label{tab:Datasets info}
% \end{table}   
% \end{footnotesize}  

\vspace{-2pt} 

\begin{table*}
% \tiny
    \centering
    \caption{Clustering performance on eight datasets (mean$\pm$std) where each experiment is performed for 10 runs. We  employ six popular metrics: accuracy, Normalized Mutual Information (NMI), Average Rand Index (ARI), and  macro F1-score are four metrics for ground-truth label analysis, whereas modularity \protect\cite{Newman_2006} and conductance \protect\cite{yang_friendster_2012} are graph-level metrics. 
    All metrics except conductance will indicate a better clustering output  with a larger value. DMAT-i results  highlighted in bold if they have the top 2 clustering performance. The asterisk indicates a convergence issue. Certain data points are missing when execution ran out of GPU memory. DGI can only handle five smaller datasets due to high GPU memory cost, and GCA also could not handle largest of these 8 datasets.}
    \label{tab: DMAT-i metrics synthetic graph}
    \begin{adjustbox}{max width=\textwidth}
    \begin{tabular}{llrrrrrrrrrr}
    \hline
    Dataset & Metric  & KMeans & DeepWalk & SDCN   & AGC   & SSGC  & AGE   &  DGI   & GCA & ProGCL  &  \bf{DMAT-i} \\
    \hline
{ACM} 
& Accuracy $\uparrow$ & $ 66.62 \pm 0.55 $ & $ 50.59 \pm 4.27 $ & $ 89.63 \pm 0.31 $  & $ 78.21 \pm 0.00 $ & $ 84.43 \pm 0.29 $  & $ 90.18 \pm 0.13 $ & $ 90.17 \pm 0.28 $ & $ 89.91 \pm 0.46 $ & $ 89.18 \pm 1.70 $ & $ \bf{91.60} \pm 0.70 $  \\
& NMI $\uparrow$ & $ 32.41 \pm 0.34 $ & $ 16.12 \pm 4.96 $ & $ 66.74 \pm 0.75 $ & $ 46.31 \pm 0.01 $ & $ 56.15 \pm 0.51 $ & $ 66.92 \pm 0.30 $ & $ 67.84 \pm 0.72 $ & $ 66.58 \pm 0.91 $ & $ 64.64 \pm 3.04 $ & $ \bf{70.95} \pm 1.44 $  \\
& ARI $\uparrow$ & $ 30.22 \pm 0.41 $ & $ 18.56 \pm 5.80 $ & $ 72.00 \pm 0.75 $ & $ 48.02 \pm 0.00 $ &  $ 60.17 \pm 0.60 $  & $ 73.12 \pm 0.31 $ & $ 73.28 \pm 0.66 $ & $ 72.49 \pm 1.08 $ & $ 70.72 \pm 3.88 $ & $ \bf{76.72} \pm 1.75 $ \\
& macro F1 $\uparrow$ &  $ 66.83 \pm 0.57 $  & $ 46.56 \pm 4.43 $  & $ 89.60 \pm 0.32 $  & $ 78.26 \pm 0.00 $  & $ 84.44 \pm 0.29 $ & $ 90.18 \pm 0.13 $ & $ 90.12 \pm 0.27 $ & $ 89.89 \pm 0.46 $ & $ 89.16 \pm 1.71 $ & $ \bf{91.59} \pm 0.70 $ \\
& Modularity $\uparrow$ & $31.20 \pm 0.50 $ & $ 38.57 \pm 9.51 $ & $60.86 \pm 0.16 $  & $ 59.44 \pm 0.02 $ & $ 60.19 \pm 0.05 $  & $ 60.93 \pm 0.08 $ & $ 59.79 \pm 0.19 $ & $ 60.05 \pm 0.12 $  & $ 60.14 \pm 0.43 $ & $ 57.92 \pm 0.20 $\\
& Conductance $\downarrow$ & $ 30.96 \pm 0.23 $ & $ 1.79 \pm 0.59 $ &  $ 3.07 \pm 0.17 $  & $ {2.51} \pm 0.01 $ & $ 2.54 \pm 0.11 $  & $ 3.64 \pm 0.19 $  & $ 3.87 \pm 0.14 $ & $ 3.85 \pm 0.18 $ & $ 4.06 \pm 0.15 $ & $ 6.68 \pm 0.27 $ \\
    \hline
{DBLP} 
& Accuracy $\uparrow$ & $ 38.65 \pm 0.58 $  & $ 38.99 \pm 0.02 $  & $ 69.08 \pm 1.95 $ & $ 69.06 \pm 0.06 $ & $ 68.66 \pm 1.95 $  & $ *62.49 \pm 0.76 $ & $ 59.72 \pm 4.68 $  & $ 77.69 \pm 0.39 $  & $ 73.79 \pm 1.70 $  & $ \bf{80.30} \pm 0.60 $  \\
& NMI $\uparrow$ & $ 11.56 \pm 0.53 $   & $ 5.91 \pm 0.02 $ & $ 34.64 \pm 1.94 $  &  $ 37.00 \pm 0.07 $ & $ 33.89 \pm 2.08 $ & $ *37.32 \pm 0.50 $ & $ 26.90 \pm 4.43 $ & $ 46.24 \pm 0.57 $ 
 & $ 41.54 \pm 1.27 $  & $ \bf{51.00} \pm 0.81 $  \\
& ARI $\uparrow$ & $ 6.95 \pm 0.39 $ & $ 5.83 \pm 0.02 $ & $ 36.31 \pm 2.86 $  & $ 33.69 \pm 0.13 $ &  $ 37.30 \pm 3.13 $  & $ *34.60 \pm 0.71 $ & $ 25.12 \pm 4.76 $ & $ 50.46 \pm 0.81 $ & $ 43.30 \pm 2.99 $ & $ \bf{55.42} \pm 1.08 $ \\
& macro F1 $\uparrow$ &  $ 31.81 \pm 0.53 $  & $ 36.87 \pm 0.02 $  & $ 67.81 \pm 3.46 $   & $ 68.59 \pm 0.05 $  & $ 65.91 \pm 2.19 $ & $ *59.16 \pm 0.83 $ & $ 59.31 \pm 4.69 $ & $ 77.29 \pm 0.37 $ & $ 72.96 \pm 2.03 $ & $ \bf{79.94} \pm 0.59 $ \\
& Modularity $\uparrow$ & $33.83 \pm 0.47 $ & $ 64.05 \pm 0.03 $ & $63.38 \pm 1.87$  &  $ {68.77} \pm 0.01 $ & $ 62.02 \pm 1.64 $  & $ *48.62 \pm 0.87 $ & $ 50.16 \pm 3.77 $ & $ 63.01 \pm 0.28 $ & $ 64.62 \pm 1.02 $  & $ 55.67 \pm 0.71 $ \\
& Conductance $\downarrow$ & $ 36.20 \pm 0.51 $ & $ 4.03 \pm 0.02 $ &  $7.56 \pm 0.54$  & $ 5.29 \pm 0.01 $  & $ {3.24} \pm 0.52 $  & $ *11.15 \pm 0.15 $  & $ 13.84 \pm 1.12 $ & $ 9.51 \pm 0.16 $ & $ 9.53 \pm 0.29 $ & $ 16.52 \pm 0.53 $ \\
    \hline    
{Cora} 
& Accuracy $\uparrow$ & $ 35.37 \pm 3.72 $  & $ 63.87 \pm 2.14 $ & $ 64.27 \pm 4.87 $  & $ 65.23 \pm 0.93 $  & $ 68.50 \pm 1.98 $   & $ 74.34 \pm 0.42 $  & $ 68.47 \pm 1.43 $   & $ 69.24 \pm 2.92 $ & $ 68.17 \pm 4.67 $  & $ \bf{70.57 \pm 1.28} $  \\
& NMI $\uparrow$ & $ 16.64 \pm 4.21 $  & $ 44.11 \pm 1.33 $ & $ 47.39 \pm 3.49 $  & $ 50.05 \pm 0.49 $  & $ {52.80} \pm 1.03 $  & $ 58.11 \pm 0.58 $ & $ 52.60 \pm 0.88 $ & $ 54.48 \pm 1.94 $ 
 & $ 54.37 \pm 2.71 $  & $ 53.59 \pm 1.22 $  \\
& ARI $\uparrow$ & $ 9.31 \pm 2.14 $ & $ 39.64 \pm 1.68 $ & $ 39.72 \pm 5.53 $ & $ 40.23 \pm 0.95 $ &  $ 45.70 \pm 1.28 $  & $ 50.87 \pm 0.96 $ & $ 45.63 \pm 1.44 $ & $ 46.63 \pm 3.25 $ & $ 45.37 \pm 5.04 $  & $ \bf{47.34 \pm 2.41} $ \\
& macro F1 $\uparrow$ & $ 31.49 \pm 4.58 $  & $ 57.98 \pm 2.43 $  & $ 57.88 \pm 6.99 $  & $ 58.93 \pm 1.68 $  & $ {64.38} \pm 2.71 $ & $ 70.37 \pm 0.29 $ & $ 65.79 \pm 1.53 $ & $ 68.10 \pm 2.68 $ & $ 67.12 \pm 4.85 $  & $ \bf{69.33 \pm 1.00} $ \\
& Modularity $\uparrow$ & $ 20.77 \pm 3.37 $ & $ 72.98 \pm 0.79 $ & $62.59 \pm 5.18 $ &  $ 69.98 \pm 0.46 $ & $ 73.71 \pm 0.45 $  & $ 71.89 \pm 0.14 $ & $ 69.86 \pm 0.29 $ & $ 74.18 \pm 0.51 $ & $ 74.36 \pm 0.38 $ & $ \bf{74.19 \pm 0.39} $ \\
& Conductance $\downarrow$ & $ 59.77 \pm 5.31 $  & $ 7.88 \pm 0.35 $ &  $ 18.32 \pm 2.26$  & $ 11.08 \pm 1.61 $  & $ 9.41 \pm 0.55 $  & $ 8.23 \pm 0.11 $ & $ 13.64 \pm 0.69 $ & $ 10.27 \pm 0.31 $ & $ 9.47 \pm 0.54 $ & $ 10.04 \pm 0.52 $ \\
    \hline  
{Citeseer} 
& Accuracy $\uparrow$ & $ 46.70 \pm 4.33 $  & $ 43.56 \pm 1.03 $  & $ 63.42 \pm 3.31 $  & $ 67.18 \pm 0.52 $    &  $ 67.86 \pm 0.26 $    & $ 66.06 \pm 0.78 $   & $ 68.68 \pm 0.76 $    & $ 66.23 \pm 1.00 $ & $ 66.43 \pm 1.16 $   & $ 67.46 \pm 0.41 $   \\
& NMI $\uparrow$ & $ 18.42 \pm 3.26 $  & $ 16.02 \pm 0.56 $ & $ 37.28 \pm 2.19 $  &  $ 41.37 \pm 0.70 $  & $ 41.86 \pm 0.22 $  & $ 40.56 \pm 0.88 $ & $ 43.22 \pm 0.91 $ & $ 40.81 \pm 1.15 $ & $ 41.41 \pm 1.03 $ & $ 41.75 \pm 0.62 $  \\
& ARI $\uparrow$ & $ 18.42 \pm 3.26 $  & $ 16.37 \pm 0.66 $  & $ 37.40 \pm 2.79 $   & $ 42.10 \pm 0.87 $  &  $ 42.95 \pm 0.30 $   & $ 39.84 \pm 0.75 $  & $ 44.53 \pm 1.02 $  & $ 41.24 \pm 1.45 $  & $ 41.73 \pm 1.52 $ & $ 42.48 \pm 0.60 $  \\
& macro F1 $\uparrow$ & $ 44.47 \pm 4.44 $  & $ 40.37 \pm 0.97 $  & $ 56.16 \pm 4.53 $   & $ 62.68 \pm 0.48 $  & $ 63.61 \pm 0.23 $ & $ 60.80 \pm 0.75 $ & $ 64.41 \pm 0.70 $ & $ 62.16 \pm 0.95 $ & $ 62.53 \pm 1.11 $ & $ 62.83 \pm 0.38 $ \\
& Modularity $\uparrow$ & $43.57 \pm 2.67 $ & $ 76.44 \pm 0.20 $ & $70.83 \pm 2.77$  & $ 77.57 \pm 0.21 $ & $ 78.03 \pm 0.12 $  & $ 71.88 \pm 0.45 $ & $ 72.42 \pm 0.38 $ & $ 73.14 \pm 0.36 $ & $ 74.54 \pm 0.44 $ & $ 75.78 \pm 0.23 $ \\
& Conductance $\downarrow$ & $ 37.21 \pm 2.19 $  & $ 2.98 \pm 0.12 $  &  $7.98 \pm 1.99 $  & $ {1.72} \pm 0.04 $  & $ 1.75 \pm 0.03 $ & $ 4.84 \pm 0.13 $ & $ 7.19 \pm 0.55 $ & $ 6.96 \pm 0.58 $ & $ 5.57 \pm 0.23 $ & $ 3.02 \pm 0.22 $ \\
    \hline    
{Pubmed} 
& Accuracy $\uparrow$ & $ 59.50 \pm 0.02 $  & $ 69.98 \pm 0.04 $   & $ 59.95 \pm 1.00 $  &   $ 61.54 \pm 0.00 $ & $ {70.71} \pm 0.00 $ & $ 69.66 \pm 0.09 $ & - & $ 64.10 \pm 2.11 $ & -  & $ \bf{70.90 \pm 0.20} $   \\
& NMI $\uparrow$ & $ 31.21 \pm 0.10 $  & $ 29.09 \pm 0.11 $ & $ 17.78 \pm 0.91 $ &   $ 29.11 \pm 0.00 $  &$ {32.12} \pm 0.00 $ & $ 29.06 \pm 0.16 $  & - & $ 28.50 \pm 2.41 $ & -  & $ \bf{32.49 \pm 0.28} $ \\
& ARI $\uparrow$ & $ 28.08 \pm 0.08 $  & $ 31.81 \pm 0.13 $   & $ 16.39 \pm 1.16 $  &   $ 26.16 \pm 0.00 $ & $ {33.26} \pm 0.00 $ & $ 31.26 \pm 0.12 $ & - & $ 26.15 \pm 2.46 $  & - & $ \bf{33.52 \pm 0.36} $ \\
& macro F1 $\uparrow$ & $ 58.15 \pm 0.02 $  & $ 68.51 \pm 0.06 $ & $ 60.29 \pm 1.02 $  & $ 60.28 \pm 0.00 $ &  $ {69.91} \pm 0.00 $ & $ 68.68 \pm 0.08 $  & - &  $ 63.69 \pm 2.34 $ & - & $ \bf{70.10 \pm 0.20} $ \\
& Modularity $\uparrow$ &  $ 34.92 \pm 0.06 $  & $ 57.25 \pm 0.26 $ & $55.53 \pm 0.86 $  & $ 50.40 \pm 0.00 $  & $ {57.73} \pm 1.35 $ & $ 57.48 \pm 0.10 $ & - & $ 53.90 \pm 1.76 $  & - & $ 57.56 \pm 0.44 $   \\
& Conductance $\downarrow$ & $ 17.27 \pm 0.04 $  & $ 4.67 \pm 0.03 $ & $ 7.50 \pm 0.58$ & $ 8.65 \pm 0.00 $ & $ {3.93} \pm 0.00 $ & $ 4.75 \pm 0.16 $ & - & $ 9.51 \pm 0.80 $ & - & $ \bf{4.10 \pm 0.20} $  \\
    \hline    
{Amazon Photo} 
& Accuracy $\uparrow$ & $ 27.86 \pm 0.81 $  & $ 77.27 \pm 2.48 $  & $ 60.42 \pm 3.36 $  & $ 55.93 \pm 0.09 $  &  $ 56.16 \pm 1.05 $  & $ 66.96 \pm 3.00 $  & $ 61.05 \pm 2.48 $ & $ 77.21 \pm 0.72 $  & $ 78.27 \pm 0.97 $  & $ 76.53 \pm 1.32 $  \\
& NMI $\uparrow$ & $ 13.78 \pm 1.19 $  & $ 68.97 \pm 1.96 $  & $ 50.08 \pm 3.28 $  &  $ 53.35 \pm 0.05 $  & $ 51.74 \pm 1.66 $ & $ 56.73 \pm 2.62 $  &  $ 52.93 \pm 2.11 $  & $ 66.48 \pm 1.23 $   & $ 70.11 \pm 1.32 $  & $ 66.94 \pm 1.37 $  \\
& ARI $\uparrow$ & $ 5.62 \pm 0.42 $  & $ 58.64 \pm 2.81 $  & $ 40.08 \pm 3.95 $ & $ 25.31 \pm 0.10 $  & $ 33.86 \pm 1.36 $ & $ 46.48 \pm 3.37 $  & $ 39.59 \pm 2.74 $  & $ 56.09 \pm 0.92 $   & $ 61.30 \pm 1.71 $ & $ 58.84 \pm 1.07 $  \\
& macro F1 $\uparrow$ & $ 23.78 \pm 0.48 $   & $ 71.59 \pm 2.47 $  & $ 53.13 \pm 5.99 $   & $ 51.56 \pm 0.06 $   & $ 52.00 \pm 0.67 $ & $ 62.13 \pm 3.33 $  & $ 59.60 \pm 2.94 $  & $ 76.23 \pm 0.71 $ & $ 72.35 \pm 1.49 $ & $ 70.05 \pm 0.77 $ \\
& Modularity $\uparrow$ & $ 8.38 \pm 0.51 $ & $ 73.18 \pm 0.12 $ & $59.25 \pm 4.04 $ & $ 57.69 \pm 0.04 $ & $ 62.07 \pm 1.88 $  & $ 64.07 \pm 1.45 $ & $ 61.12 \pm 1.39 $ & $ 67.76 \pm 0.83 $ & $ 70.81 \pm 0.47 $ & $ 70.72 \pm 0.19 $ \\
& Conductance $\downarrow$ & $ 76.38 \pm 0.58 $  & $ 8.47 \pm 0.23 $  &  $20.17 \pm 3.88$  & $ {4.42} \pm 0.00 $   & $ 8.37 \pm 2.15 $ & $ 15.81 \pm 1.49 $ & $ 22.14 \pm 1.65 $ & $ 15.27 \pm 1.11 $ & $ 10.46 \pm 0.72 $ & $ 10.98 \pm 0.67 $ \\
    \hline      
{Coauthor CS} 
& Accuracy $\uparrow$ & $ 27.96 \pm 1.09 $  & $ 67.10 \pm 2.98 $  & $ 56.86 \pm 3.40 $  & $ 62.24 \pm 1.81 $  & $ 66.19 \pm 1.19 $ & $ 76.35 \pm 3.14 $ & - & $ 72.02 \pm 2.54 $ & -  & $ \bf{76.92 \pm 1.26} $  \\
& NMI $\uparrow$ & $ 15.42 \pm 2.25 $ & $ 66.67 \pm 0.86 $  & $ 54.79 \pm 2.44 $    & $ 65.22 \pm 0.44 $ & $ 70.06 \pm 0.67 $  &  $ 76.75 \pm 1.66 $   & - & $ 73.95 \pm 1.02 $ & -  & $ 72.55 \pm 0.41 $  \\
& ARI $\uparrow$ & $ 1.02 \pm 0.74 $  & $ 53.66 \pm 2.91 $  & $ 40.41 \pm 4.52 $  &   $ 46.96 \pm 3.54 $ & $ 58.50 \pm 0.17 $ & $ 71.27 \pm 5.46 $ & - & $ 63.92 \pm 3.21 $ & -  & $ 66.91 \pm 1.38 $ \\
& macro F1 $\uparrow$ & $ 11.68 \pm 1.56 $ & $ 63.36 \pm 2.84 $ & $ 29.36 \pm 3.22 $  & $ 51.42 \pm 1.27 $ & $ {60.17} \pm 1.94 $ & $ 71.10 \pm 1.96 $  & - &  $ 63.63 \pm 3.28 $ & - & $ \bf{70.48 \pm 3.13} $ \\
& Modularity $\uparrow$ &  $ 9.61 \pm 1.88 $  & $ 72.88 \pm 0.41 $ & $53.05 \pm 2.02$  & $ 69.58 \pm 0.14 $  & $ 71.82 \pm 0.14 $ & $ 70.45 \pm 1.71 $ & - & $ 69.91 \pm 0.58 $ & - & $ 69.79 \pm 0.40 $   \\
& Conductance $\downarrow$ & $ 37.12 \pm 4.10 $  & $ 17.09 \pm 0.66 $ & *$23.09 \pm 1.89 $ & $ 19.80 \pm 0.24 $ & $ 19.76 \pm 0.22 $  & $ 14.41 \pm 0.32 $ & - & $ 21.96 \pm 0.60 $  & -  
 & $ 21.13 \pm 0.66 $  \\
    \hline   
{Coauthor PHY} 
& Accuracy $\uparrow$ & $ 56.19 \pm 0.75 $  & $ 87.97 \pm 0.01 $  & $ 64.65 \pm 6.92 $  & $ {77.41} \pm 0.00 $  & $ 55.70 \pm 2.26 $ & $ 92.04 \pm 0.06 $ & - & - & - &  $ \bf{89.30 \pm 0.70} $ \\
& NMI $\uparrow$ & $ 11.72 \pm 1.92 $ & $ 69.13 \pm 0.02 $   & $ 50.60 \pm 3.71 $ & $ 62.11 \pm 0.02 $ & $ 57.71 \pm 1.31 $  & $ 75.84 \pm 0.13 $  & - & - & - & $ \bf{72.54} \pm 0.80 $ \\
& ARI $\uparrow$ & $ 8.25 \pm 1.26 $  & $ 79.15 \pm 0.03 $  & $ 48.76 \pm 9.58 $  & $ {72.43} \pm 0.02 $  & $ 44.91 \pm 1.58 $ & $ 84.44 \pm 0.16 $ & - & - & - & $ \bf{77.68} \pm 1.61 $ \\
& macro F1 $\uparrow$ & $ 24.74 \pm 2.11 $ & $ 83.32 \pm 0.02 $   & $ 48.51 \pm 4.68 $   & $ 62.09 \pm 0.00 $   & $ 55.26 \pm 2.30 $ & $ 88.90 \pm 0.08 $ & - & - & - &  $ \bf{86.65} \pm 0.91 $ \\
& Modularity $\uparrow$ & $ 5.74 \pm 0.83 $  & $ 47.96 \pm 0.00 $  & $44.97 \pm 3.16$   & $ 45.31 \pm 0.00 $  & $ {60.70} \pm 0.39 $ & $ 47.69 \pm 0.07 $ & -  & - & - & $ 50.56 \pm 0.27 $ \\
& Conductance $\downarrow$ & $ 10.56 \pm 1.47 $  & $ 5.99 \pm 0.00 $  & $19.86 \pm 7.16$  & $ {5.80} \pm 0.00 $  & $ 13.47 \pm 0.07 $ & $ 5.73 \pm 0.03 $ & - & - & - & $ 7.31 \pm 0.31 $ \\
    \hline      
    \end{tabular} 
    \end{adjustbox}
    % }
\end{table*} 

%\vspace*{-3cm}

% \begin{table*}
%     \small
%     \centering
%     \caption{Clustering performance for ogb-arxiv.}
%     \begin{adjustbox}{max width=\textwidth}
%     \begin{tabular}{lrrrrrr}
%     \hline
%     Method & Accuracy & NMI & ARI &  macro F1 $\uparrow$ & Modularity $\uparrow$ &  Conductance $\downarrow$ \\
%     \hline
%     DMAT-i & $ 27.50 \pm 0.22 $ & $ 39.76 \pm 0.15 $ & $ 16.21 \pm 0.35 $ &  $ 21.67 \pm 0.37 $ & $ 38.02 \pm 0.48 $ & $ 58.14 \pm 0.50 $  \\
%     SSGC & $ 36.54 \pm 1.50 $ & $ 44.83 \pm 0.30 $ & $ 30.24 \pm 1.82 $ &  $ 23.80 \pm 0.87 $ & $ 62.08 \pm 0.26 $ & $ 28.39 \pm 0.90 $  \\
%     \hline
%     \end{tabular} 
%     \end{adjustbox}
%     \label{tab: ogb-arxiv}
% \end{table*}

\vspace{-2pt} 
\begin{table*}
    \small
    \centering
    \caption{Accuracy for semi-supervised node classification task with different data usage for embedding generation: 1) using 10\% of data with labels; 2) using all data without labels.}
    \begin{adjustbox}{max width=\textwidth}
    \begin{tabular}{llrrrrrrrr}
    \hline
     Data Usage & Method & Cora & Citeseer & Pubmed & ACM  & DBLP & Amazon Photo & Coauthor CS & Coauthor PHY \\
    \hline
    train data (labeled)  & DMT  & $ 84.30 \pm 0.25 $ & $ 70.42 \pm 0.33 $ & $ 86.46 \pm 0.16 $ & $ 91.42 \pm 0.36 $ & $ 77.59 \pm 0.30 $ & $ 92.60 \pm 0.42 $ & $ 93.30 \pm 0.12 $ & $ 95.44 \pm 0.03 $ \\
    & DMAT  & $ 83.92 \pm 0.45 $ & $ 71.39 \pm 0.38 $ & $ 86.19 \pm 0.10 $ & $ 92.04 \pm 0.16 $ & $ 79.80 \pm 0.60 $ & $ 93.42 \pm 0.11 $ & $ 93.44 \pm 0.15 $ & $ 95.20 \pm 0.04 $ \\
    & DMAT-i  & $ 81.99 \pm 0.54 $ & $ 70.91 \pm 0.27 $ & $ 83.52 \pm 0.21 $ & $ 91.32 \pm 0.38 $ & $ 78.39 \pm 0.67 $ & $ 93.19 \pm 0.19 $ & $ 92.90 \pm 0.12 $ & $ 94.86 \pm 0.07 $ \\
    \hline
    all data (unlabeled) & DMAT-i  & $ 83.65 \pm 0.71 $ & $ 72.40 \pm 0.43 $ & $ 83.91 \pm 0.25 $ & $ 92.55 \pm 0.40 $ & $ 80.92 \pm 0.50 $ & $ 92.97 \pm 0.16 $ & $ 91.28 \pm 0.17 $  & $ 94.66 \pm 0.08 $ \\
    & SSGC  & $ 83.48 \pm 0.06 $ & $ 68.15 \pm 0.02 $ & $ 84.59 \pm 0.01 $ & $ 89.71 \pm 0.25 $ & $ 77.14 \pm 0.12 $ & $ 89.80 \pm 0.14 $ & $ 91.37 \pm 0.03 $  & $ 94.88 \pm 0.02 $ \\
    & GCA  & $ 83.89 \pm 0.56 $ & $ 73.36 \pm 0.34 $ & $ 83.38 \pm 0.17 $ & $ 90.01 \pm 0.27 $ & $ 79.73 \pm 0.50 $ & $ 90.30 \pm 0.47 $ & $ 90.91 \pm 0.11 $  &  -  \\
    & ProGCL  & $ 85.04 \pm 0.42 $ & $ 71.42 \pm 0.39 $ & - & $ 88.98 \pm 0.48 $ & $ 79.55 \pm 0.41 $ & $ 92.13 \pm 0.82 $ &  -  &  -  \\
    & AGE  & $ 83.78 \pm 0.22 $ & $ 72.13 \pm 0.92 $ & $ 80.18 \pm 0.24 $ & $ 92.10 \pm 0.18 $ & $ 80.02 \pm 0.40 $ & $ 73.16 \pm 2.53 $ & $ 91.40 \pm 0.13 $  & $  94.21 \pm 0.08 $ \\
    \hline
    \end{tabular} 
    \end{adjustbox}
    % }
    \label{tab: node classificatoin metrics}
\end{table*}

\paragraph{Baselines:}
For the node clustering task, we compared the proposed DMAT-i model with multiple frameworks: 1) \textbf{KMeans} \cite{Kmeans_1979} (when applied to attributed graphs  uses 
node attributes only);  
2) \textbf{DeepWalk} \cite{Perozzi_deepwalk_2014},  which uses topological information only, and seven recent 
frameworks that leverage both node attributes and graph 
structure: 
3) \textbf{AGC (2019)}\cite{ijcai2019_AGC}  that  uses high-order graph convolution; 4) \textbf{DGI (2019)}~\cite{velickovic_2019_DGI} maximizes mutual information between patch representations and high-level summaries of graph; 5) \textbf{SDCN (2020)} \cite{Bo_2020_structural_deep} 
 unifies an autoencoder module with a GCN module;  6) \textbf{AGE (2020)}~\cite{Cui_AGE_2020} applies a customized Laplacian smoothing filter; 7) \textbf{SSGC (2021)} \cite{ssgc_2021}  is a variant of GCN that  exploits a modified Markov Diffusion Kernel.  8) \textbf{GCA (2021)}~\cite{zhu2021_GCA} leverages a node-level contrastive loss between two augmented graph views to learn a graph representation; 9) \textbf{ProGCL (2022)}~\cite{xia2022_progcl}, on top of GCA, further proposed a more suitable measure for negatives hardness and similarity. 
%  (In the appendix, we 
% mention some of the other candidate baselines, and explain why they were not chosen). 
% It is noteworthy that AGC, SSGC, AGE are using
% Laplacian smoothing filters, which DMAT-i is also using but in a scalable fashion.
To compare performance on node classification and link prediction, we select the most competitive graph embedding based frameworks correspondingly.

% : AGE, GCA, and SSGC.  AGE and SSGC employ Lapalcian filtering while GCA utilizes contrastive mechanism, and they were  
% most competitive with respect to  clustering. Among others, several like 
% K-means and SDCN are dedicated for node clustering only and 
% thus not applicable to these tasks. 

% \paragraph{Parameter Settings:} Our DM(A)T framework is implemented in PyTorch 1.7 on CUDA 10.1, whereas   the graph filtering procedure is in C++. Our experiments are performed on nodes with a dual Intel Xeon 6148s @2.4GHz CPU and dual NVIDIA Volta V100 w/ 16GB memory GPU and 384 GB DDR4 memory. 
% Graph filtering is executed on CPU while tuplet loss based training process is performed on a single GPU. We applied a AdamW optimization method with a decoupled weight decay regularization technique \cite{loshchilov2019decoupled_weight_decay} (see  
% appendix). 

\begin{figure}
    \centering
    \includegraphics[width=0.7\columnwidth]{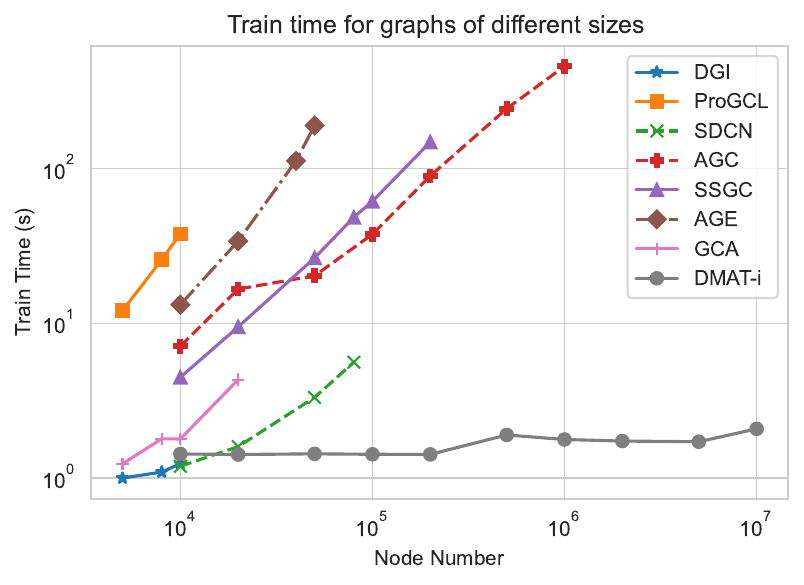}
    \caption{Scalability of Different Frameworks: Training Time vs. No. of Nodes in Graph }
    \label{fig:Time_VS_Size}
\end{figure}

\vspace*{-2ex} 
\paragraph{Scalability of Representation Construction}  
%Synthetic datasets were used to evaluate scalability of DMAT-i and other  frameworks.
% We used PaRMAT \cite{PaPMAT_2015} to generate undirected synthetic graphs of growing size with edge count set as 20 times the number of nodes and a random feature matrix with a dimension of 1000. For each clustering method, we performed 5-epoch training, repeated  each experiment 5 times, and report average times.  
As in Figure~\ref{fig:Time_VS_Size}, all baselines hit specific ceilings as limited by the GPU memory capacity while DMAT-i can continuously scale with application of mini-batch training and use of random-walk to obtain approximate pagerank scores.  
Particularly, DMAT-i could handle $10^7$ nodes, and no other frameworks could handle more than $10^6$ nodes.
The details of experimental settings are in appendix.

\vspace*{-4ex} 

\vspace*{-2ex} 
\subsection{Results on  Downstream Tasks} 
\vspace*{-2ex} 

DM(A)T is evaluated on performance of semi-supervised node classification while DMAT-i is evaluated for multiple tasks: node clustering, node classification, 
and link prediction. Our framework is compared with existing state-of-the-art appraoches on 8 real-world datasets (with details in appendix). 
%{\bf I will suggest moving these details in rest of this paragraph to appendix.}

\vspace*{-2ex} 
\paragraph{Node Clustering}
We set the number of clusters to the number of ground-truth classes and perform K-Means algorithm \cite{Kmeans_1979} on resulting embedding $Z$ from DMAT-i 
following previous efforts~\cite{GALA_park_2019,Bo_2020_structural_deep,ssgc_2021,Cui_AGE_2020}.  
Table \ref{tab: DMAT-i metrics synthetic graph} summarizes clustering results.  
DMAT-i maintains either the state-of-the-art clustering results or is  fairly close to the best. In particular, DMAT-i further reduced state-of-the-art accuracy gap between unsupervised learning and transductive supervised learning as presented later in Table \ref{tab: node classificatoin metrics} across  datasets such as DBLP and Coauthor PHY. 
Not surprisingly,  deep clustering methods that use both node attributes and graph structure appear to be more robust and stronger than those  using either of 
them (KMeans and DeepWalk), although  the latter  shows good performance for certain datasets. Compared with GCN based methods like SDCN, DMAT-i shows significant performance gain due to solving over-smoothing issues through graph filtering. For clustering methods like AGC, SSGC or AGE with carefully designed Laplacian-smoothing filters, DMAT-i can still outperform them in most cases. The most competitive clustering performance comes from AGE on several datasets -- however, it does not  even converge for DBLP.
DMAT-i achieves robust convergence across all real-word datasets  --  a detailed summary of convergence time across different datasets is presented in the appendix.

% DMAT-i achieves robust convergence across all real-word datasets  --  detailed comparison of accuracy and training time for different methods on this dataset is shown in Figure~\ref{fig:Acc VS Time}. GCA and DGI are not shown because they could not even execute with this dataset.  Overall, we can see that DMAT-i obtains best  accuracy, while converging in much less time compared with the most competitive baseline AGE.    
% A detailed summary of convergence time across different datasets is presented in Table \ref{tab: train time table}. One observation we can make is that GCA becomes several times more expensive that DMAT-i for all the  datasets with 15,000 or more edges (and cannot be executed for Coauthor PHY). AGE, as noted above, 
% is significantly more expensive, whereas other methods that are faster do not produce results of the same quality. 

\vspace*{-2ex} 

\begin{table}
    \small
    \centering
    \caption{Link prediction performance.}
    \begin{adjustbox}{max width=0.7\columnwidth}
    \begin{tabular}{llrrrrr}
    \hline
    Dataset & Metrics & DMAT-i & SSGC & AGE & GCA & ProGCL \\
    \hline
    Cora  
    & AP  & $ 92.41 \pm 0.28 $ & $ 93.24 \pm 0.00 $ & $ 92.26 \pm 0.30 $ & $ 92.95 \pm 0.41 $  & $ 92.87 \pm 0.28 $ \\
      & AUC  & $ 92.62 \pm 0.29 $ & $ 92.14 \pm 0.00 $ & $ 92.07 \pm 0.21 $ & $ 92.95 \pm 0.34 $  & $ 93.60 \pm 0.13 $ \\
    Citeseer 
    & AP  & $ 95.52 \pm 0.26 $ & $ 96.14 \pm 0.00 $ & $ 92.22 \pm 0.48 $ & $ 93.38 \pm 0.39 $ & $ 95.65 \pm 0.28 $ \\
    & AUC  & $ 95.19 \pm 0.26 $ & $ 95.29 \pm 0.00 $ & $ 92.66 \pm 0.44 $ & $ 92.57 \pm 0.49 $ & $ 95.59 \pm 0.24 $ \\
    Pubmed  
    & AP & $ 95.42 \pm 0.08 $ & $ 97.53 \pm 0.00 $ & $ 84.67 \pm 0.09 $ & $ 92.65 \pm 0.44 $  & - \\
    & AUC  & $ 95.18 \pm 0.10 $ & $ 97.84 \pm 0.00 $ & $ 86.70 \pm 0.12 $ & $ 93.81 \pm 0.37 $  & - \\
    ACM  
    & AP & $ 97.55 \pm 0.17 $ & $ 82.33 \pm 0.00 $ & $ 98.14 \pm 0.10 $ & $ 92.61 \pm 1.05 $ & $ 97.02 \pm 0.23 $ \\
    & AUC  & $ 97.41 \pm 0.17 $ & $ 81.15 \pm 0.00 $ & $ 97.51 \pm 0.18 $ & $ 94.19 \pm 0.75 $ & $ 97.31 \pm 0.18 $ \\
    DBLP  
    & AP & $ 95.50 \pm 0.37 $ & $ 95.88 \pm 0.00 $ & $ 92.68 \pm 0.31 $ & $ 93.41 \pm 0.61 $   & $ 95.99 \pm 0.24 $ \\
    & AUC  & $ 95.50 \pm 0.50 $ & $ 95.22 \pm 0.00 $ & $ 90.96 \pm 0.43 $ & $ 92.47 \pm 0.59 $ & $ 95.38 \pm 0.23 $ \\
    Amazon Photo 
    & AP & $ 92.73 \pm 0.23 $ & $ 83.93 \pm 0.00 $ & $ 91.65 \pm 0.23 $  & $ 76.26 \pm 1.39 $ & $ 93.43 \pm 0.65 $ \\
    & AUC  & $ 93.89 \pm 0.20 $ & $ 89.21 \pm 0.00 $ & $ 93.12 \pm 0.19 $ & $ 81.28 \pm 1.34 $  & $ 95.66 \pm 0.42 $ \\
    Coauthor CS
    & AP  & $ 94.76 \pm 0.14 $ & $ 88.96 \pm 0.00 $ & $ 93.96 \pm 0.18 $ & $ 82.70 \pm 0.98 $  & - \\
    & AUC  & $ 95.03 \pm 0.12 $ & $ 93.56 \pm 0.00 $ & $ 93.61 \pm 0.15 $ & $ 83.54 \pm 0.74 $  & - \\
    Coauthor PHY  
    & AP & $ 91.25 \pm 0.20 $ & $ 93.92 \pm 0.00 $ & $ 94.35 \pm 0.08 $ & - & - \\
    & AUC  & $ 92.75 \pm 0.15 $ & $ 96.57 \pm 0.00 $ & $ 95.20 \pm 0.06 $  & - & - \\
    \hline
    \end{tabular} 
    \end{adjustbox}
    % }
    \label{tab: link prediction metrics}
\end{table}

% \begin{small}
% \begin{table}[tbh]
%     % \centering
%     \small
%     \caption{DMAT-i empirical evaluation of $\tau^{0}$ distribution with several quantiles: Q1 (25\%) and Q3 (75\%)}
%     \begin{adjustbox}{max width=0.7\columnwidth,center}
%     \begin{tabular}{lrrrrrrrr}
%     \hline
%     Quantile  & Coauthor CS & Coauthor PHY & Amazon Photo &	ACM & Citeseer & DBLP &	Cora & Pubmed \\ 
%     \hline
%     Q1  & $ 1.90 $  & $ 1.48 $  & $ 1.26 $   & $ 5.76 $ & $ 1.91 $ & $ 2.38 $ & $ 1.06 $ & $ 3.52 $ \\
%     Median & $ 5.11 $  & $ 6.21 $  & $ 2.74 $ & $ 11.67 $ & $ 4.09 $ & $ 4.50 $ & $ 2.32 $ & $ 8.09 $ \\
%     Q3  & $ 16.18 $  & $ 16.35 $  & $ 5.70 $ & $ 25.47 $ & $ 9.14 $ & $ 9.47 $ & $ 5.88 $ & $ 18.37 $ \\
%     \hline
%     \end{tabular}
%     \end{adjustbox}
%     \label{tab: tau dist}
% \end{table}   
% \end{small} 

% \vspace*{-2ex} 

\paragraph{Node Classification }
For the transductive semi-supervised node classification task, we applied train-validation-test data split with fraction as train (10\%),  validation (10\%),  and test (80\%). Following the experimental settings of SSGC \cite{ssgc_2021} and GCA \cite{zhu2021_GCA}, we evaluate the classification performance of DM(A)T and DMAT-i by using a linear classifier to perform semi-supervised classification and report the accuracy. 
As shown in Table \ref{tab: node classificatoin metrics}, the embedding generation methods are categorized based on the availability of labels, where DM(A)T learns from train data (10\% of all data) with labels and DMAT-i can proceed in an  unsupervised way on all data samples. For comparison, we apply DMAT-i in both settings. 

With labelled training data, DM(A)T turns out to achieve high quality of representations and shows superior results. DMAT-i, however, fails to recognize part of positive samples as compared with DMAT in this condition and lose some accuracy. 
When labels are completely unavailable, we can see that competitive results have been observed from DMAT-i compared to other advanced baselines under unsupervised setting.
More importantly, DMAT-i generally achieves better performance when generating embedding in unsupervised condition than ``partially-supervised'' condition (with partial labels available). That is because much more samples i.e. all data, are included during tuplet loss optimization.

\vspace*{-2ex} 
\paragraph{Link Prediction}

To evaluate   DMAT-i on this task, 
 we remove 5\% edges for validation and 10\% edges for test while keeping all node attributes~\cite{kipf2016variational_VGAE,GALA_park_2019,Cui_AGE_2020}. 
The reconstructed adjacency matrix $\hat{A}$ can be calculated as  per the previous publication~\cite{kipf2016variational_VGAE}: $\hat{A}$=$\sigma(ZZ^{T})$, where $\sigma$ denotes the sigmoid function. For comparison purposes,  we report area under the ROC curve (AUC) and average precision (AP) following settings from previous works \cite{kipf2016variational_VGAE,GALA_park_2019,Cui_AGE_2020}.  As shown in Table \ref{tab: link prediction metrics}, DMAT-i  is robust, i.e,  produces high-quality link prediction (above 90\% for both metrics for all datasets), whereas no other  methods has  a comparable consistency.  

% \paragraph{Empirical evaluation of $\tau^{0}$}
% Compared with the ideal contrastive loss~\cite{chuang_2020_debiased}, DMAT-i introduces false negative samples whose effect is covered in $\tau^{0}$.  Based on Eq.\ref{eq: tau0}, we empirically evaluate $\mathbb{E}_{\substack{x^- \sim p_x^-} } h(x,x^-)$ by $\widehat{\mathbb{E}}_{x^{-} \in \mathcal{X}}h(x,x^-) $, i.e. the average of all negative samples of each sample $x$ across all samples $\mathcal{X}$. Similar empirical approximation will be applied to $\mathbb{E}_{\substack{x^+ \sim p_x^+} } h(x,x^+)$. Most $\tau^{0}$ values from most samples remain reasonable small values as in
% Table~\ref{tab: tau dist}. 

\vspace*{-2ex} 
\section{Conclusions} 
\vspace*{-2ex} 

This paper  has presented  a  scalable  graph  (node-level) 
learning framework.  Employing a mutli-class tuplet loss  function, we have introduced both semi-supervised learning and unsupervised  algorithms.  
We have also established connections between tuplet loss and  contrastive loss functions and also theoretically shown how our method leads to generalization error bound on the downstream classification task.
The learned representation is used for three downstream   tasks: node clustering,  classification, and link prediction. Our extensive evaluation has shown  better scalability over any existing method, and consistently high accuracy (state-of-the-art  or very competitive in each case).

% \input{input/related_work}

% \newtheorem{innercustomlemma}{Lemma}
% \newenvironment{customlemma}[1]
%   {\renewcommand\theinnercustomlemma{#1}\innercustomlemma}
%   {\endinnercustomlemma}
  
% \newtheorem{innercustomthm}{Theorem}
% \newenvironment{customthm}[1]
%   {\renewcommand\theinnercustomthm{#1}\innercustomthm}
%   {\endinnercustomthm}

% \onecolumn
\newcommand{\sjf}[1]{\textcolor{blue}{SJ: #1}}

% \newpage
\appendix

\section{Proofs for Theoretical Analysis}

\subsection{Assumptions and Notations}

The underlying set of discrete latent classes are denoted as $\mathcal{C}(\cdot)$ representing semantic content and the distribution over classes as $\rho(y)$,i.e. probability of observing class $y$. Given a pair of sample vectors $(x, x^{\prime})$ from $\mathcal{X} \in \mathbb{R}^{N \times d}$, $p_x^+(x^\prime) = p(x^\prime | C(x^\prime) = C(x))$ is the probability of observing $x^\prime$ as a positive sample for $x$ and $p_x^-(x^\prime) = p(x^\prime | C(x^\prime) \neq C(x))$ the probability of a negative one. Assume that the class probabilities $\rho(y) = \tau^+$ are uniform, i.e. the fraction of each class in each tuplet stays constant. $\tau^+$ of each class can be empirically calculated as the fraction of each class in the whole dataset. The data distribution can be decomposed as: 

\[ p(x^\prime) = \tau^+p_x^+(x^\prime) + \tau^- p_x^-(x^\prime) \]

We define $h(x, x^\prime;f)  = exp\{\frac{f(x)^{T} \cdot f(x^\prime)}{t}\}$. For simplicity of analysis, we assume $t = 1$ yielding $h(x, x^\prime;f)  = exp\{f(x)^{T} \cdot f(x^\prime)\}$

\subsection{Proof of Lemma \ref{lemma: app DMAT inequality}}

\begin{customlemma}{1}
\label{lemma: app DMAT inequality}
For any embedding $ f $, given the same size of tuplets sharing one positive sample ${x_{0}^{+}}$, i.e. $(x, x_{0}^+,\{x^-_i\}_{i=1}^{N-1})$ for $L_{\textnormal{Unbiased}}^{\mathcal{N}+1} $ and $(x, x_{0}^{+}, \{x^+_i\}_{i=1}^m,  \{x^-_i\}_{i=1}^q )$ for $   L_{\textnormal{DM(A)T}}^{\textnormal{m,q}} $, we have:

    \begin{align*}
        L_{\textnormal{DM(A)T}}^{\textnormal{m,q}}(f) 
        \leq  \widetilde L_{\textnormal{Unbiased}}^{\mathcal{N}+1}(f) 
        % \leq L_{\substack{\textnormal{DMAT-i}}}^{\textnormal{m,q}}(f)
    \end{align*}
\end{customlemma}

% We first focus on $L_{\substack{\textnormal{DMAT}}}^{\textnormal{m,q}}(f)$ for simplicity since $L_{\substack{\textnormal{DMT}}}^{\textnormal{m,q}}(f)$ and $L_{\substack{\textnormal{DMAT}}}^{\textnormal{m,q}}(f)$  share the similar form.

\begin{proof}[Proof of Lemma \ref{lemma: app DMAT inequality}] 

We first focus on the DMAT loss. Consider the DMAT tuplet \\
$(x, x_{0}^{+}, \{x^+_i\}_{i=1}^m,  \{x^-_i\}_{i=1}^q )$ with a fixed size ($\mathcal{N}+1$), we have $\mathcal{N}+1 = 2 + m + q $, i.e. $\mathcal{N}-1 =  m + q $

$L_{\textnormal{DMAT}}^{\textnormal{m,q}}(f)$ is the empirical estimate of $\widetilde L_{\textnormal{DMAT}}^{\textnormal{m,q}}(f)$ as follows:

\begin{align*}
    \widetilde L_{\textnormal{DMAT}}^{\textnormal{m,q}}(f) =  -\log \frac{h(x, x_{0}^{+})+ m\mathbb{E}_{x^+ \sim p_x^+} h(x, x^+)}{h(x, x_{0}^{+}) + m\mathbb{E}_{x^+ \sim p_x^+} h(x, x^+) + q\mathbb{E}_{x^- \sim p_x^-} h(x, x^-)}
\end{align*}

In comparison with $\widetilde L_{\textnormal{Unbiased}}^{\mathcal{N}+1}$ ~\cite{chuang_2020_debiased} on the tuplet $(x, x_{0}^{+},\{x^-_i\}_{i=1}^{N-1})$:

\begin{align*}
    \widetilde L_{\textnormal{Unbiased}}^{\mathcal{N}+1}(f) = -\log \frac{h(x, x_{0}^{+})}{h(x, x_{0}^{+}) + (\mathcal{N}-1)\mathbb{E}_{x^- \sim P_x^-} h(x, x^-)} \\
    = -\log \frac{h(x, x_{0}^{+})}{h(x, x_{0}^{+}) + (m+q)\mathbb{E}_{x^- \sim P_x^-} h(x, x^-)}
\end{align*}

Then we have:

\begin{align*}
        \widetilde L_{\textnormal{DMAT}}^{\textnormal{m,q}}(f)
        \leq 
        -\log \frac{h(x, x_{0}^{+})}{h(x, x_{0}^{+}) + q\mathbb{E}_{x^- \sim P_x^-} h(x, x^-)}
        \leq  \widetilde L_{\textnormal{Unbiased}}^{\mathcal{N}+1}(f)
\end{align*}

The first inequality is based on the fact that $\frac{a+c}{b+c} \geq \frac{a}{b}$ for $a\leq b$ and $a,b,c\geq 0$ and the second inequality is due to $m \geq 0$.

For $L_{\textnormal{DMT}}^{\textnormal{m,q}}(f)$ on the tuplet $(x, x_{0}^{+}, \{x^+_i\}_{i=1}^m,  \{x^-_i\}_{i=1}^q )$,  $L_{\textnormal{DMT}}^{\textnormal{m,q}}(f)$ is the empirical estimate of $\widetilde L_{\textnormal{DMT}}^{\textnormal{m,q}}(f)$ as follows:

\begin{align*}
    \widetilde L_{\textnormal{DMT}}^{\textnormal{m,q}}(f) =  -\log \frac{h(x, x) + h(x, x_{0}^{+})+ m\mathbb{E}_{x^+ \sim p_x^+} h(x, x^+)}{h(x, x) + h(x, x_{0}^{+}) + m\mathbb{E}_{x^+ \sim p_x^+} h(x, x^+) + q\mathbb{E}_{x^- \sim p_x^-} h(x, x^-)}
\end{align*}

The only difference between  $\widetilde L_{\textnormal{DMT}}^{\textnormal{m,q}}(f)$ and  $\widetilde L_{\textnormal{DMAT}}^{\textnormal{m,q}}(f)$ is that  $\widetilde L_{\textnormal{DMT}}^{\textnormal{m,q}}(f)$ includes an additional term $h(x,x)$. All the proof above for $L^{m,q}_{\textnormal{DMAT}}$ still holds for $L^{m,q}_{\textnormal{DMT}}$.

\end{proof}

\subsection{Proof of Theorem \ref{thm: app bound objective diff}}

We follow a similar proof strategy in~\cite{chuang_2020_debiased}. 
To prove Theorem 1, we first seek a bound on the tail probability that the difference between the integrands of two objective functions $\widetilde L_{\textnormal{DMAT-i}}^{\textnormal{m,q}}(f)$ and $\widetilde L_{\textnormal{Unbiased}}^{\mathcal{N}+1}(f)$ given the same size of tuplets is greater than a specific threshold $\varepsilon$:

\begin{align*}
    \mathbb{P}(\Delta \geq \varepsilon), \quad  \Delta 
    &=  \bigg|  L_{\textnormal{DMAT-i}}^{\textnormal{m,q}}(f) - \widetilde L_{\textnormal{Unbiased}}^{\mathcal{N}+1}(f) \bigg | \\
    &= \bigg | -\log \frac{ h(x,\bar{x})}{ h(x,\bar{x}) + \sum_{i=1}^m  h(x, x^+_i) + \sum_{i=1}^q  h(x, x^-_i)} \\
    & + \log \frac{ h(x,\bar{x})}{ h(x,\bar{x}) + (m+q) \mathbb{E}_{x^- \sim p_x^-}  h(x,x^-)} \bigg |
\end{align*}

where $\Delta$ depends on $x, \bar{x}$ and the collections of samples $\{x^+_i\}_{i=1}^m $ and $ \{x^-_i\}_{i=1}^q$. Here we apply $\mathcal{N}-1 =  m + q $ as from \textit{Proof of Lemma \ref{lemma: app DMAT inequality}}. This tail will be controlled by the following lemma.

\begin{customlemma}{A.1}\label{lemma: bound Delta}

With $x$ and $\bar{x}$ in $\widetilde X$ fixed, let $\{x^+_i\}_{i=1}^m$ and $ \{x^-_i\}_{i=1}^q$ be collections of i.i.d. random variables sampled from $p_x^+$ and $p_x^-$ respectively. Then $\forall \varepsilon > 0$, 

\begin{align*}
\mathbb{P}(\Delta \geq \varepsilon) \leq 2 \exp \left ( - \frac{m \varepsilon^2}{2(e^3-e)(\tau^{0})^{2}} \right ) 
+ 2 \exp \left ( - \frac{q \varepsilon^2}{2(e^3-e)(\tau^{-})^{2}} \right ) \end{align*}

where 
\vspace{-5pt}

\begin{align*}
\tau^{0} = \tau^{+}  \bigg ( \frac{\big | \frac{1}{m} \sum_{i=1}^m  h(x, x^+_i)  - \mathbb{E}_{\substack{x^- \sim p_x^-} } h(x,x^-) \big |}{\big | \frac{1}{m} \sum_{i=1}^m  h(x, x^+_i)  - \mathbb{E}_{\substack{x^+ \sim p_x^+} } h(x,x^+) \big |} \bigg )
\end{align*}

\end{customlemma}

\begin{proof}[Proof of Lemma \ref{lemma: bound Delta}] 

First, we define $g(x, \{x^+_i\}_{i=1}^m, \{x^-_i\}_{i=1}^q)$ as: 

\begin{align*}
g(x, \{x^+_i\}_{i=1}^m, \{x^-_i\}_{i=1}^q)=\frac{1}{m+q} \left( \sum_{i=1}^m  h(x, x^+_i) + \sum_{i=1}^q  h(x, x^-_i) \right) 
\end{align*}

\begin{align*}
\mathbb{P}(\Delta \geq \varepsilon) 
&=\mathbb{P} \bigg ( \bigg | \log \big \{ h(x,\bar{x}) + (m+q)g(x, \{x^+_i\}_{i=1}^m, \{x^-_i\}_{i=1}^q) \big \} \\
& - \log \big \{ h(x,\bar{x}) + (m+q) \mathbb{E}_{x^- \sim p_x^-}  h(x,x^-) \big \} \bigg | \geq \varepsilon \bigg )  \\
&=\mathbb{P} \bigg (   \log \big \{ h(x,\bar{x}) + (m+q)g(x, \{x^+_i\}_{i=1}^m, \{x^-_i\}_{i=1}^q) \big \} \\
& - \log \big \{ h(x,\bar{x}) + (m+q) \mathbb{E}_{x^- \sim p_x^-}  h(x,x^-) \big \}   \geq \varepsilon \bigg ) \\
&\quad + \mathbb{P} \bigg ( - \log \big \{ h(x,\bar{x}) + (m+q)g(x, \{x^+_i\}_{i=1}^m, \{x^-_i\}_{i=1}^q) \big \} \\
 &+ \log \big \{ h(x,\bar{x}) + (m+q) \mathbb{E}_{x^- \sim p_x^-}  h(x,x^-) \big \}   \geq \varepsilon \bigg ) \\
& = \mathbb{P}_{1}(\varepsilon) + \mathbb{P}_{2}(\varepsilon)
\end{align*}

The first term can be bounded as:

\begin{align}
 \mathbb{P}_{1}(\varepsilon)
 &= \mathbb{P} \bigg (  \log \frac{ h(x,\bar{x}) + (m+q)g(x, \{x^+_i\}_{i=1}^m, \{x^-_i\}_{i=1}^q)  }{  h(x,\bar{x}) + (m+q) \mathbb{E}_{x^- \sim p_x^-}  h(x,x^-)  }  \geq \varepsilon \bigg ) \nonumber
    \\
    &\leq \mathbb{P} \bigg (\frac{(m+q)g(x, \{x^+_i\}_{i=1}^m, \{x^-_i\}_{i=1}^q) - (m+q) \mathbb{E}_{x^- \sim p_x^-} h(x,x^-) }{ h(x,\bar{x}) + (m+q) \mathbb{E}_{x^- \sim p_x^-}  h(x,x^-)} \geq \varepsilon \bigg ) \nonumber
    \\
    & = \mathbb{P} \bigg  (g(x, \{x^+_i\}_{i=1}^m, \{x^-_i\}_{i=1}^q) - \mathbb{E}_{x^- \sim p_x^-} h(x,x^-)   \geq \\
    & \varepsilon \bigg \{ \frac{1}{m+q} h(x,\bar{x}) +  \mathbb{E}_{x^- \sim p_x^-}  h(x,x^-) \bigg \} \bigg ) \nonumber
    \\ 
    &\leq \mathbb{P} \bigg (g(x, \{x^+_i\}_{i=1}^m, \{x^-_i\}_{i=1}^q) - \mathbb{E}_{x^- \sim p_x^-} h(x,x^-)  \geq \varepsilon e^{-1}\bigg ).\label{a_eq_1}
\end{align}

The first inequality follows by applying the fact that  $\log x \leq x -1 $ for $x > 0$. The second inequality holds since $ \frac{1}{m+q} h(x,\bar{x}) +  \mathbb{E}_{x^- \sim p_x^-}  h(x,x^-)  \geq e^{-1}$. Similarly, the second term can be bounded as:

\begin{align}
     \mathbb{P}_{2}(\varepsilon)
    &= \mathbb{P} \bigg ( \log \frac{ h(x,\bar{x}) + (m+q) \mathbb{E}_{x^- \sim p_x^-}  h(x,x^-)}{ h(x,\bar{x}) + (m+q)g(x, \{x^+_i\}_{i=1}^m, \{x^-_i\}_{i=1}^q)} \geq \varepsilon \bigg ) \nonumber
    \\
    &\leq \mathbb{P} \bigg (\frac{ (m+q) \mathbb{E}_{x^- \sim p_x^-} h(x,x^-) - (m+q)g(x, \{x^+_i\}_{i=1}^m, \{x^-_i\}_{i=1}^q) }{h(x,\bar{x})+  (m+q)g(x, \{x^+_i\}_{i=1}^m, \{x^-_i\}_{i=1}^q)} \geq \varepsilon \bigg ) \nonumber
    \\
    & = \mathbb{P} \bigg  (  \mathbb{E}_{x^- \sim p_x^-} h(x,x^-)  - g(x, \{x^+_i\}_{i=1}^m, \{x^-_i\}_{i=1}^q) \geq \\ 
    & \quad \varepsilon  \bigg \{  \frac{1}{m+q}h(x,\bar{x})  +  g(x, \{x^+_i\}_{i=1}^m, \{x^-_i\}_{i=1}^q) \bigg \}  \bigg ) \nonumber
    \\
    &\leq \mathbb{P} \bigg (  \mathbb{E}_{x^- \sim p_x^-} h(x,x^-) - g(x, \{x^+_i\}_{i=1}^m, \{x^-_i\}_{i=1}^q)  \geq \varepsilon e^{-1} \bigg ) . \label{a_eq_2}
\end{align} 

Combining Eq.\eqref{a_eq_1} and Eq.\eqref{a_eq_2}, we have
\begin{align*}
\mathbb{P}(\Delta \geq \varepsilon) \leq \mathbb{P} \bigg ( \big |g(x, \{x^+_i\}_{i=1}^m, \{x^-_i\}_{i=1}^q) - \mathbb{E}_{x^- \sim p_x^-} h(x,x^-) \big |  \geq \varepsilon e^{-1} \bigg )
\end{align*}

\begin{align*}
    &\big |g(x, \{x^+_i\}_{i=1}^m, \{x^-_i\}_{i=1}^q) - \mathbb{E}_{x^- \sim p_x^-} h(x,x^-) \big | \\
    &\;\;\; = \bigg | \frac{1}{m+q} \left( \sum_{i=1}^m  h(x, x^+_i) + \sum_{i=1}^q  h(x, x^-_i) \right)  - \mathbb{E}_{\substack{x^- \sim p_x^-} } h(x,x^-) \bigg | \\
    &\;\;\; = \bigg | \frac{m}{m+q}\frac{1}{m} \sum_{i=1}^m  h(x, x^+_i) + \frac{q}{m+q}\frac{1}{q} \sum_{i=1}^q  h(x, x^-_i)   - (\tau^{+} + \tau^{-}) \mathbb{E}_{\substack{x^- \sim p_x^-} } h(x,x^-) \bigg | \\
    &\;\;\; = \bigg |\tau^{+} \bigg (\frac{1}{m} \sum_{i=1}^m  h(x, x^+_i)  - \mathbb{E}_{\substack{x^- \sim p_x^-} } h(x,x^-) \bigg) 
    + \tau^{-} \bigg ( \frac{1}{q} \sum_{i=1}^q  h(x, x^-_i)   -  \mathbb{E}_{\substack{x^- \sim p_x^-} } h(x,x^-) \bigg ) \bigg | \\
    &\;\;\; \leq \tau^{+} \bigg | \frac{1}{m} \sum_{i=1}^m  h(x, x^+_i)  - \mathbb{E}_{\substack{x^- \sim p_x^-} } h(x,x^-) \bigg |
    + \tau^{-} \bigg | \frac{1}{q} \sum_{i=1}^q  h(x, x^-_i)   -  \mathbb{E}_{\substack{x^- \sim p_x^-} } h(x,x^-) \bigg | \\
    &\;\;\; = \tau^{0} \bigg | \frac{1}{m} \sum_{i=1}^m  h(x, x^+_i)  - \mathbb{E}_{\substack{x^+ \sim p_x^+} } h(x,x^+) \bigg |
    + \tau^{-} \bigg | \frac{1}{q} \sum_{i=1}^q  h(x, x^-_i)   -  \mathbb{E}_{\substack{x^- \sim p_x^-} } h(x,x^-) \bigg | \\
\end{align*}

where 
\vspace{-5pt}

\begin{align*}
\tau^{0} = \tau^{+}  \bigg ( \frac{\big | \frac{1}{m} \sum_{i=1}^m  h(x, x^+_i)  - \mathbb{E}_{\substack{x^- \sim p_x^-} } h(x,x^-) \big |}{\big | \frac{1}{m} \sum_{i=1}^m  h(x, x^+_i)  - \mathbb{E}_{\substack{x^+ \sim p_x^+} } h(x,x^+) \big |} \bigg )
\end{align*}

Here $\frac{m}{m+q} \simeq \tau^+$ and $\frac{q}{m+q} \simeq \tau^-$ are based on the assumption of uniform class distribution, i.e. the fraction of each class stays constant in each tuplet: $\rho(y) = \tau^+$. Further, we have:

\begin{align*}
&\mathbb{P} \bigg ( \big |g(x, \{x^+_i\}_{i=1}^m, \{x^-_i\}_{i=1}^q) - \mathbb{E}_{x^- \sim p_x^-} h(x,x^-) \big |  \geq \varepsilon e^{-1} \bigg ) \\
&\leq \mathbb{P} \bigg ( \tau^{0} \bigg | \frac{1}{m} \sum_{i=1}^m  h(x, x^+_i)  - \mathbb{E}_{\substack{x^+ \sim p_x^+} } h(x,x^+) \bigg | \\
 & \quad + \tau^{-} \bigg | \frac{1}{q} \sum_{i=1}^q  h(x, x^-_i)   -  \mathbb{E}_{\substack{x^- \sim p_x^-} } h(x,x^-) \bigg |  \geq \varepsilon e^{-1} \bigg ) \\
&\leq \text{I} (\varepsilon) + \text{II} (\varepsilon).
\end{align*}

where 
\vspace{-5pt}
\begin{align*}
&\text{I} (\varepsilon)=  \mathbb{P} \left ( \tau^{0} \bigg | \frac{1}{m} \sum_{i=1}^m  h(x, x^+_i)  - \mathbb{E}_{\substack{x^- \sim p_x^+} } h(x,x^-) \bigg |  \geq \frac{\varepsilon e^{-1}}{2} \right )  \\
&\text{II}(\varepsilon)=  \mathbb{P} \left ( \tau^{-} \bigg | \frac{1}{q} \sum_{i=1}^q  h(x, x^-_i)   -  \mathbb{E}_{\substack{x^- \sim p_x^-} } h(x,x^-) \bigg | \geq \frac{\varepsilon e^{-1}}{2} \right ).
\end{align*}

Hoeffding's inequality states that if  $X, X_1, \ldots , X_N$ are i.i.d random variables bounded in the range $[a,b]$:
\vspace{-5pt}

\begin{align*}
    \mathbb{p} \left(  \big | \frac{1}{n}\sum_{i=1}^N X_i  - \mathbb{E}X \big | \geq    \varepsilon \right) \leq 2\exp \left(-\frac{2N \varepsilon^2}{b-a } \right)
\end{align*}

With $e^{-1} \leq h (x,\bar{x}) \leq e$ in our case, then:

\begin{align*}
\text{I} (\varepsilon)   \leq  2 \exp \left ( - \frac{m \varepsilon^2}{2(e^3-e)(\tau^{0})^{2}} \right ) \quad \text{and} \quad \text{II}(\varepsilon)  \leq  2 \exp \left ( - \frac{q \varepsilon^2}{2(e^3-e)(\tau^{-})^{2}} \right )
\end{align*}
\end{proof}

% \begin{proof}
% We have

% \begin{align*}
%     \{|X| + |Y| \geq 2\varepsilon^{2} \}  \subset \{ |X| \geq \varepsilon^{2} \} \cup     \{ |Y| \geq \varepsilon^{2} \}
% \end{align*}

% which yields

% \begin{align*}
%  \mathbb{P}(|X| + |Y| \geq 2\varepsilon^{2})  \leq \mathbb{P}(\{ |X| \geq \epsilon^{2} \} \cup \{ |Y| \geq \epsilon^{2} \})
%  \leq \mathbb{P}(|X| \geq \epsilon^{2}) + \mathbb{P}(|Y| \geq \epsilon^{2})
% \end{align*}

% \end{proof}

With Lemma~\ref{lemma: bound Delta} at hand, we are ready to prove Theorem~\ref{thm: app bound objective diff}.

\begin{customthm}{1}
\label{thm: app bound objective diff}
For any embedding $f$ and same size of tuplets,  we have:
\begin{align}
    \left | \widetilde L_{\textnormal{Unbiased}}^{\mathcal{N}+1}(f) - L_{\substack{\textnormal{DMAT-i}}}^{\textnormal{m,q}}(f) \right |
    \leq \sqrt{ \frac{2(e^3-e)(\tau^{0})^{2} \pi}{m}} +   \sqrt{ \frac{2(e^3-e)(\tau^{-})^{2} \pi}{q}} \nonumber 
\end{align}
\end{customthm}

\begin{proof}

By Jensen's inequality, we can push the absolute value inside the expectation to see that $ |\widetilde L_{\textnormal{Unbiased}}^{\mathcal{N}+1}(f) - L_{\substack{\textnormal{DMAT-i}}}^{\textnormal{m,q}}(f)| \leq \mathbb{E}\Delta$. Further, we write the expectation of $\Delta$ for fixed $x,\bar{x}$  as the integral of its tail probability,

\begin{align*}
\mathbb{E}\ \Delta = \mathbb{E}_{x,\bar{x}}  & \left [  \mathbb{E} [ \Delta | x,\bar{x}] \right ]  = \mathbb{E}_{x,\bar{x}} \left [ \int _0 ^\infty \mathbb{P}(\Delta \geq \varepsilon | x,\bar{x}) \text{d} \varepsilon \right ]  \\
&\leq \int _0 ^\infty   2 \exp \left ( - \frac{m \varepsilon^2}{2(e^3-e)(\tau^{0})^{2}} \right ) \text{d} \varepsilon  + \int _0 ^\infty   2 \exp \left ( - \frac{q \varepsilon^2}{2(e^3-e)(\tau^{-})^{2}} \right )  \text{d} \varepsilon \\
& = \sqrt{ \frac{2(e^3-e)(\tau^{0})^{2} \pi}{m}} +   \sqrt{ \frac{2(e^3-e)(\tau^{-})^{2} \pi}{q}} 
\end{align*}

Since the tail probably bound of Theorem \ref{lemma: bound Delta} holds uniformly for all fixed $x,\bar{x}$, the outer expectation can be removed~\cite{chuang_2020_debiased}. Both integrals in the final step can be computed via:

\vspace{-5pt}
 \[ \int_0^\infty e^{-c z^2} \text{d}z = \frac{1}{2}\sqrt{ \frac{\pi}{c}}.\] 
\vspace{-5pt}
 
\end{proof}

\subsection{Proof of Theorem~\ref{thm: app bound sup diff}}

Theorem~\ref{thm: app bound sup diff} aims to derive a data dependent bound from $L_{\substack{\textnormal{DMAT-i}}}^{\textnormal{m,q}}(f)$ on the downstream supervised generalization error $L_{\textnormal{Sup}}(f)$. To prove Theorem~\ref{thm: app bound sup diff}, we need to employ Lemma~\ref{lemma: empirical risk}, Lemma~\ref{lemma: lipschitz} and Lemma~\ref{lemma: unbiased and sup} provided as follows.

For one batch of samples $\mathcal{X}_B$ of size ${\cal{B}}$ together with their augmented counterparts, we have a $2{\cal{B}}$-tuplet  $(x, \bar{x}, \{x^+_i\}_{i=1}^m,  \{x^-_i\}_{i=1}^q )$ with $\bar{x}$ as the augmented counterpart (a trivial positive sample) of $x$. $\{x^+_i\}$ are $m$ positive samples other than $\bar{x}$ and $\{x^-_i\}$ are $q$ negative samples. Rewrite $L_{\textnormal{DMAT-i}}^{m,q}(f) $ as:

\begin{align*}
   L_{\textnormal{DMAT-i}}^{m,q}(f) 
  & =  - \log  \left \{ \frac{e^{f(x)^\top f(\bar{x})} } { e^{f(x)^\top f(\bar{x})} + \sum_{i=1}^{m}e^{f(x)^\top f(x^{+}_{i})} + \sum_{i=1}^{q}e^{f(x)^\top f(x^{-}_{i})} } \right \} \\ 
  & = \log \left \{  1 + \sum_{i=1}^{m}e^{f(x)^\top (f(x^{+}_{i})-f(\bar{x}))} + \sum_{i=1}^{q}e^{f(x)^\top (f(x^{-}_{i})-f(\bar{x}))} \right \} \\
  & = \ell \left (  \left \{ f(x)^\top \big ( f(x^+_i) - f(\bar{x}) \big  ) \right \}_{i=1}^m , \left  \{ f(x)^\top \big ( f(x^-_i) - f(\bar{x}) \big  ) \right \}_{i=1}^q \right ) \\
\end{align*}

For simplicity, we denote the loss as:

\begin{align*}
  \ell( \{ a_i \}_{i=1}^m ,  \{ b_i \}_{i=1}^q ) 
  &=  \log \left \{  1 + \sum_{i=1}^{m}a_{i} + \sum_{i=1}^{q}b_{i} \right \}.
 \end{align*}

where 
\vspace{-5pt}

\begin{align*}
a_i = e^{f(x)^\top (f(x^{+}_{i})-f(\bar{x}))} \quad and \quad b_i = e^{f(x)^\top (f(x^{-}_{i})-f(\bar{x}))} 
\end{align*}

To derive our bound, we will exploit  a concentration of measure result due to \cite{arora2019theoretical}. They consider an unsupervised loss of the form 

\[ L_{un}(f) = \mathbb{E} \left [ \ell (  \{ f(x)^\top \big (  f(x^-_i) - f(x^+) \big  )  \}_{i=1}^k )\right ], \]

where $(x,x^+, x^-_1, \ldots , x^-_k)$  are sampled from any fixed distribution on $\mathcal{X}^{k+2}$ with $x^+, x^-_i$ representing positive and negative sample of $x$ respectively (they were particularly focused on the case where $x^-_i \sim p$, but the proof holds for arbitrary distributions~\cite{chuang_2020_debiased}).  
Let $\mathbb{F}$ be a class of representation functions $\mathcal{X} \rightarrow \mathbb{R}^d$ such that $\| f(\cdot)\| \leq R$ for $R>0$. The corresponding empirical risk minimizer is:  

 \[ \hat{f} \in \arg \min_{f \in \mathbb{F} } \, \frac{1}{N} \sum_{j=1}^N \ell \left (  \{ f(x_j)^\top \big (  f(x_{ji}^-) - f(x_j^+) \big  )  \}_{i=1}^k \right ) \]
 
over a training set $\mathcal{S}=\{(x_j,x_j^+,x_{j1}^-,\dots,x_{jk}^-)\}_{j=1}^N$ of i.i.d. samples. Their result bounds the loss of the empirical risk minimizer as follows.

\begin{customlemma}{A.2}\label{lemma: empirical risk} \textnormal{\cite{arora2019theoretical}}
Let $\ell:\mathbb{R}^k\rightarrow\mathbb{R}$ be $\eta$-Lipschitz and bounded by $\Gamma$. Then with probability at least $1-\delta$ over the training set $\mathcal{S}=\{(x_j,x_j^+,x_{j1}^-,\dots,x_{jk}^-)\}_{j=1}^N$, for all $f\in \mathbb{F}$

\begin{equation}
L_{un}(\hat{f})\leq  L_{un}(f)+\mathcal{O} \left(\frac{\eta R \sqrt{k} \mathcal{R}_\mathcal{S}(\mathbb{F})}{N}+  \Gamma\sqrt{\frac{\log{\frac{1}{\delta}}}{N}} \right) \nonumber 
\end{equation}

where

\begin{equation}
\mathcal{R}_\mathcal{S}(\mathbb{F})=\mathbb{E}_{\sigma \sim \{\pm1\}^{(k+2)dN}} \left[ \sup_{f\in \mathbb{F}} \langle \sigma, f_{|\mathcal{S}} \rangle \right], \nonumber 
\end{equation}

and  $f_{|\mathcal{S}}=\left(f_t(x_j),f_t(x_j^+),f_t(x_{j1}^-),\dots, ,f_t(x_{jk}^-)\right)_{\substack{j\in[N] \\ t \in [d]}} $.

\end{customlemma}

In our particular case, $f$ are normalized embeddings with $\| f(\cdot)\| \leq 1$ and thus $e^{-1} \leq a_i, b_i \leq e $. We have $k = m+q $ and $R = 1$. So, it remains to obtain constants $\eta$ and $\Gamma$ such that $ \ell( \{ a_i \}_{i=1}^m ,  \{ b_i \}_{i=1}^q )$ is $\eta$-Lipschitz, and bounded by $\Gamma$. 

\begin{customlemma}{A.3}\label{lemma: lipschitz}
With  $e^{-1} \leq a_i, b_i \leq e $, the function $ \ell( \{ a_i \}_{i=1}^m ,  \{ b_i \}_{i=1}^q )$ is $\eta$-Lipschitz, and bounded by $\Gamma$ for 
\[ \eta =  \frac{e\sqrt{m+q}}{m+q + e} , \quad \quad \quad \Gamma = \mathcal{O} \left ( \log (m+q) \right ). \] 
\end{customlemma}

\begin{proof}
First, it is easily observed that $\ell$ is upper bounded by plugging in $a_i = b_i = e$ yielding a bound of
\begin{align*}
 \log \left \{  1 +  \sum_{i=1}^{m}e + \sum_{i=1}^{q}e \right \} = \mathcal{O} \left ( \log (m+q) \right )
 \end{align*}

To bound the Lipschitz constant we view $\ell$ as a composition  $ \ell( \{ a_i \}_{i=1}^m ,  \{ b_i \}_{i=1}^q ) = \phi \left ( g  \left(  \{ a_i \}_{i=1}^m ,  \{ b_i \}_{i=1}^q \right ) \right ) $
 
\begin{align*}
&\phi(z) = \log \left ( 1 + z \right )  \\
& z = g( \{a_i\}_{i=1}^m, \{b_i\}_{i=1}^q) = \sum_{i=1}^{m}a_{i} + \sum_{i=1}^{q}b_{i} 
\end{align*}
 
where $ (m+q)e^{-1} \leq z \leq (m+q)e$ and thus $\partial_z \phi(z) = \frac{1}{ 1 + z} \leq  \frac{e}{m+q + e}$. We therefore conclude that $\phi$ is $\frac{e}{m+q + e}$-Lipschitz. 
The Lipschitz constant of $g$ is bounded by the Forbenius norm of the Jacobian of $g$, which equals \\
$\sqrt{\sum_{i=1}^{m}(\frac{\partial{g}}{\partial{a_i}})^{2} + \sum_{i=1}^{q}(\frac{\partial{g}}{\partial{b_i}})^{2} } = \sqrt{m+q} $ 

\end{proof}

\begin{customlemma}{A.4}\label{lemma: unbiased and sup}
~\cite{chuang_2020_debiased}
For any embedding $f$ for a downstream $K$-way classification, whenever $N \geq K$ we have
\[ L_{\textnormal{Sup}}(f)  \leq \widetilde L_{\textnormal{Unbiased}}^{N+1}(f).\]
\end{customlemma}

Here, we change notation of $L_{\textnormal{Unbiased}}^{N}(f)$ from~\cite{chuang_2020_debiased} into $L_{\textnormal{Unbiased}}^{N+1}(f)$ for consistency in our work. The number of negative samples is $N-1$, requiring $N-1 \geq K-1$ as in~\cite{chuang_2020_debiased} yields $N \geq K$. Now we have all the requested lemmas, together with Theorem~\ref{thm: app bound objective diff} we are ready to prove Theorem~\ref{thm: app bound sup diff} as follows.

\begin{customthm}{2}
~\label{thm: app bound sup diff}
With probability at least $1-\delta$, for all $f \in \mathbb{F}$ and $q \geq K-1$,

 \begin{align}
      L_{\textnormal{Sup}}(\hat{f})  \leq  L_{\substack{\textnormal{DMAT-i}}}^{\textnormal{m,q}}(f) + \mathcal{O} \left (\tau^{0} \sqrt{ \frac{1}{m}} + \tau^{-} \sqrt{\frac{1}{q}} + 
      \frac{\lambda \mathcal{R}_\mathcal{S}(\mathbb{F})}{N}+  \Gamma\sqrt{\frac{\log{\frac{1}{\delta}}}{N}} \right ) \nonumber 
 \end{align}
 
where $\lambda = \eta \sqrt{k} = \frac{(m+q)e}{m + q + e}$ 
and $\Gamma = \log (m+q) $.

\end{customthm}

\begin{proof}
By Lemma~\ref{lemma: unbiased and sup}  and Theorem~\ref{thm: app bound objective diff}, requiring the number of negative samples $q \geq K-1$, we have 

\begin{align*}
 L_{\textnormal{sup}}(\hat{f}) \leq  L_{\substack{\textnormal{DMAT-i}}}^{\textnormal{m,q}}(\hat{f}) + \sqrt{ \frac{2(e^3-e)(\tau^{0})^{2} \pi}{m}} + \sqrt{ \frac{2(e^3-e)(\tau^{-})^{2} \pi}{q}}
\end{align*}

As shown earlier, our unsupervised loss $L_{\substack{\textnormal{DMAT-i}}}^{\textnormal{m,q}}(f)$ follows the same form of $L_{un}(f)$ as in Lemma~\ref{lemma: empirical risk}. Combining Lemma~\ref{lemma: empirical risk} and Lemma~\ref{lemma: lipschitz},  with probability at least $1-\delta$, for all $f \in \mathbb{F}$, we have:

\begin{align*}
    L_{\substack{\textnormal{DMAT-i}}}^{\textnormal{m,q}}(\hat f) \leq L_{\substack{\textnormal{DMAT-i}}}^{\textnormal{m,q}}(f) + \mathcal{O} \left(\frac{\lambda \mathcal{R}_\mathcal{S}(\mathbb{F})}{N}+  \Gamma\sqrt{\frac{\log{\frac{1}{\delta}}}{N}} \right),
\end{align*}
where $\lambda = \eta \sqrt{k} = \frac{(m+q)e}{m + q + e}$ 
and $\Gamma = \log (m+q) $.

\end{proof}

\clearpage

\section{Additional Results}

\subsection{T-SNE from DMAT-i node clustering on Cora Dataset}

\begin{figure*}
     \centering
     \begin{subfigure}[b]{0.32\textwidth}
         \centering
         \includegraphics[width=\textwidth]{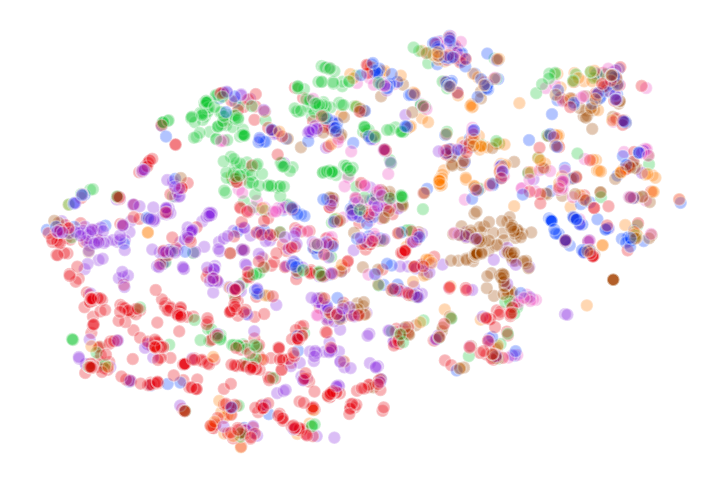}
         \caption{Raw node attributes}
     \end{subfigure}
     \hfill
     \begin{subfigure}[b]{0.32\textwidth}
         \centering
         \includegraphics[width=\textwidth]{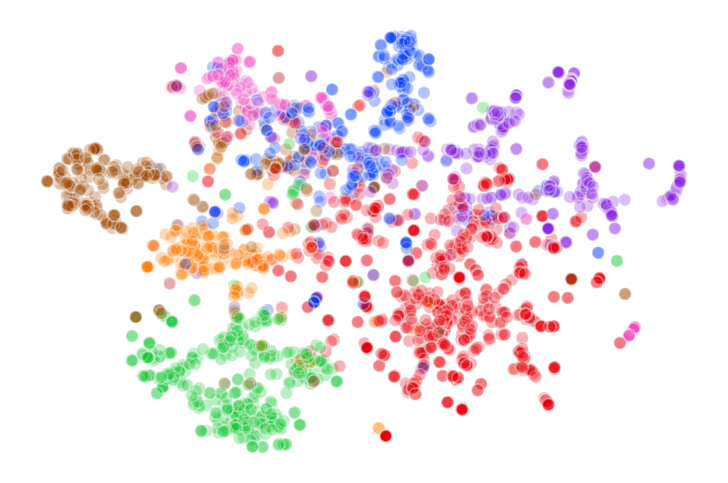}
         \caption{Smoothed attributes}
     \end{subfigure}
     \hfill
     \begin{subfigure}[b]{0.32\textwidth}
         \centering
         \includegraphics[width=\textwidth]{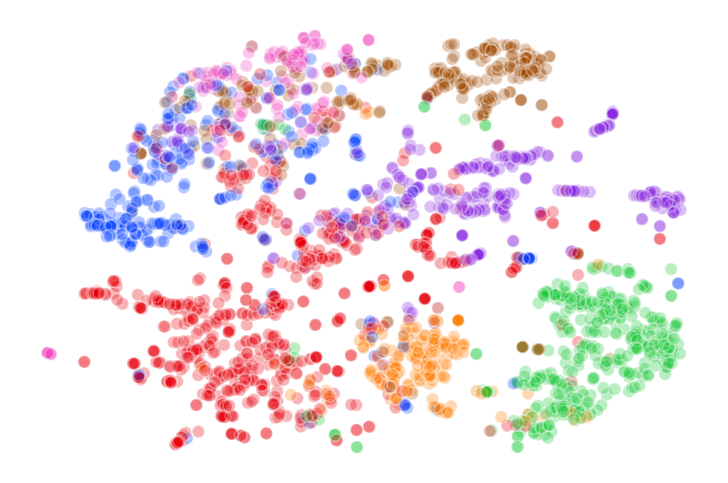}
         \caption{Encoder}
     \end{subfigure}
    \caption{Cora t-SNE for different embeddings. Each color represents a distinct class.}
    \label{fig: cora tsne embedding model}
\end{figure*}

Figure~\ref{fig: cora tsne embedding model} intuitively shows comparison between node embeddings using t-SNE algorithm \cite{t-SNE_algo}. We can see that the graph filter and encoder module can gradually recognize node semantic distribution with less overlapping areas.

\subsection{DMAT-i Training Process and Convergence Time}

\begin{figure*}
     \centering
     \begin{subfigure}[b]{0.5\textwidth}
         \centering
         \includegraphics[width=\textwidth]{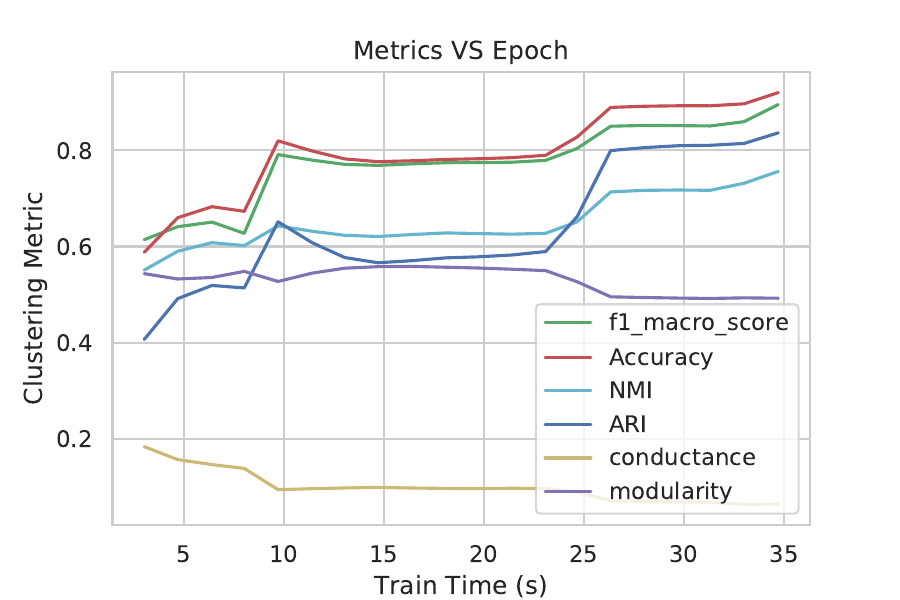}
         \caption{Clustering Metrics Convergence Curve}
         \label{fig: Coauthor_PHY Clustering metrics}
     \end{subfigure}
     \hfill
     \begin{subfigure}[b]{0.45\textwidth}
         \centering
         \includegraphics[width=\textwidth]{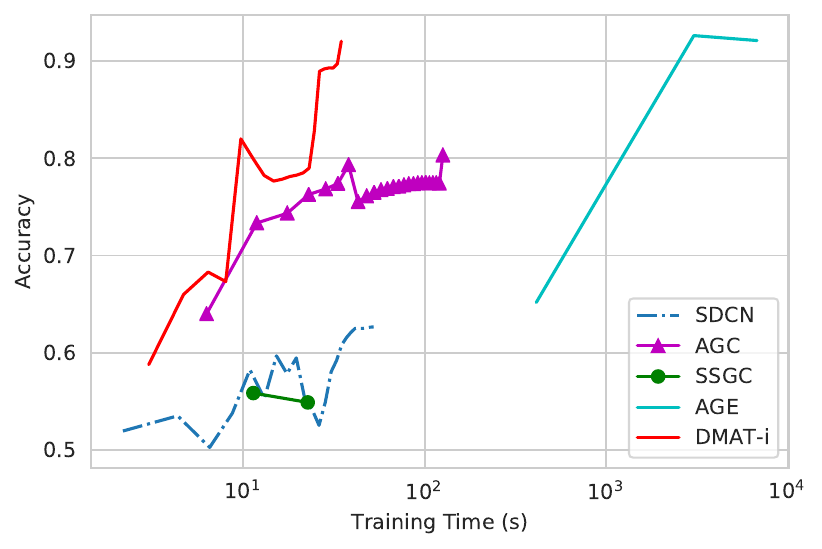}
         \caption{Accuracy Vs. Training Time}
         \label{fig:Acc VS Time}
     \end{subfigure}
    \caption{DMAT-i Training Process on Coauthor PHY}
    \label{fig:Coauthor PHY Training}
\end{figure*}

\begin{table*}
    \centering
    \caption{Convergence time (s) comparison. The asterisk indicates a convergence issue with stop time reported.}
    \begin{adjustbox}{max width=\textwidth}
    \begin{tabular}{lrrrrrrrr}
    \hline
    Dataset & ACM  & DBLP & Cora  & Citeseer & Pubmed & Coauthor PHY & Coauthor CS & Amazon Photo \\
    \hline
    SDCN  & $ 3.74 \pm 0.52 $  & $ 3.37 \pm 0.52 $  & $ 6.52 \pm 0.73 $ & $ 5.18 \pm 1.16 $  & $23.29 \pm 0.62 $ & $ 52.20 \pm 0.15 $ & $ 22.78 \pm 0.70 $ & $ 11.67 \pm 0.84 $ \\
    ProGCL   & $ 29.87 \pm 2.50 $  & $ 65.99 \pm 1.48 $  & $ 17.99 \pm 1.42 $ & $ 35.72 \pm 1.72 $  & -  & -  & - & $ 293.36 \pm 16.79 $ \\
    AGC   & $ 1.75 \pm 0.20 $  & $ 0.57 \pm 0.22 $  & $ 0.82 \pm 0.28 $ & $ 7.49 \pm 1.10 $  & $ 0.94 \pm 0.22 $  & $ 125.44 \pm 0.83 $ & $ 17.78 \pm 0.64 $ & $ 3.43 \pm 0.57 $ \\
    SSGC   & $ 5.83 \pm 0.94 $  & $ 1.92 \pm 0.74 $  & $ 2.18 \pm 1.88 $ & $ 22.24 \pm 3.87 $  & $ 26.29 \pm 0.31 $  & $ 23.65 \pm 0.73 $ & $ 37.31 \pm 14.69 $ & $ 6.53 \pm 3.12 $ \\
    AGE   & $ 95.61 \pm 0.16 $  & *$ 88.57 \pm 0.80 $  & $ 70.36 \pm 0.34 $ & $ 182.10 \pm 0.52 $  & $ 1343.35 \pm 40.15 $  & $ 6799.57 \pm 33.19 $ & $ 463.10 \pm 1.89 $ & $ 87.15 \pm 1.11 $ \\
    DGI   & $ 15.01 \pm 0.70 $  & $ 16.18 \pm 0.63 $  & $ 13.64 \pm 0.91 $ & $ 15.22 \pm 0.62 $  & -  & - & - & $ 13.12 \pm 0.45 $ \\
    GCA   & $ 8.09 \pm 0.64 $  & $ 11.27 \pm 2.34 $  & $ 7.43 \pm 0.57 $ & $ 8.65 \pm 0.56 $  & $ 74.76 \pm 2.81 $  & - & $ 144.71 \pm 0.74 $ & $ 47.24 \pm 3.06 $ \\
    \bf{DMAT-i}   & $ 10.15 \pm 0.85 $  & $ 16.04 \pm 0.70 $  & $ 3.06 \pm 0.71 $ & $ 14.25 \pm 1.13 $  & $ 3.26 \pm 0.37 $  & $ 33.86 \pm 1.35 $  & $ 37.41 \pm 1.99 $  & $ 17.49 \pm 2.71 $ \\
    \hline
    \end{tabular} 
    \end{adjustbox}
    % }
    \label{tab: train time table}
\end{table*}

DMAT-i achieves robust convergence across all real-word datasets  -- Figure \ref{fig: Coauthor_PHY Clustering metrics} shows the training process of DMAT-i on Coauthor PHY with all metrics converged to a steady state and detailed comparison of accuracy and training time for different methods on this dataset is shown in Figure~\ref{fig:Acc VS Time}. GCA and DGI are not shown because they could not even execute with this dataset.  Overall, we can see that DMAT-i obtains best  accuracy, while converging in much less time compared with the most competitive baseline AGE.    

A detailed summary of convergence time across different datasets is presented in Table \ref{tab: train time table}. One observation we can make is that GCA becomes several times more expensive than DMAT-i for all the  datasets with 15,000 or more edges (and cannot be executed for Coauthor PHY). AGE is significantly more expensive, whereas other methods that are faster do not produce results of the same quality.

\subsection{Empirical evaluation of $\tau^{0}$}

\begin{figure*}
     \centering
     \begin{subfigure}[b]{0.32\textwidth}
         \centering
         \includegraphics[width=\textwidth]{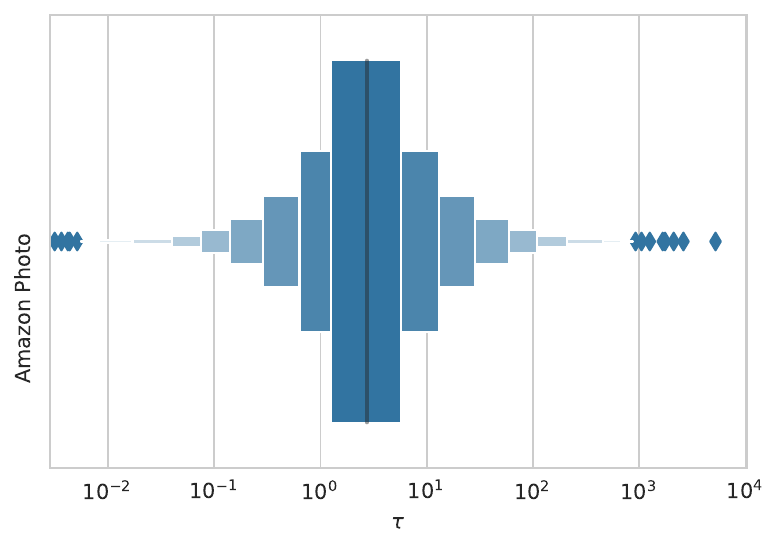}
         \caption{Amazon Photo}
     \end{subfigure}
     \hfill
     \begin{subfigure}[b]{0.32\textwidth}
         \centering
         \includegraphics[width=\textwidth]{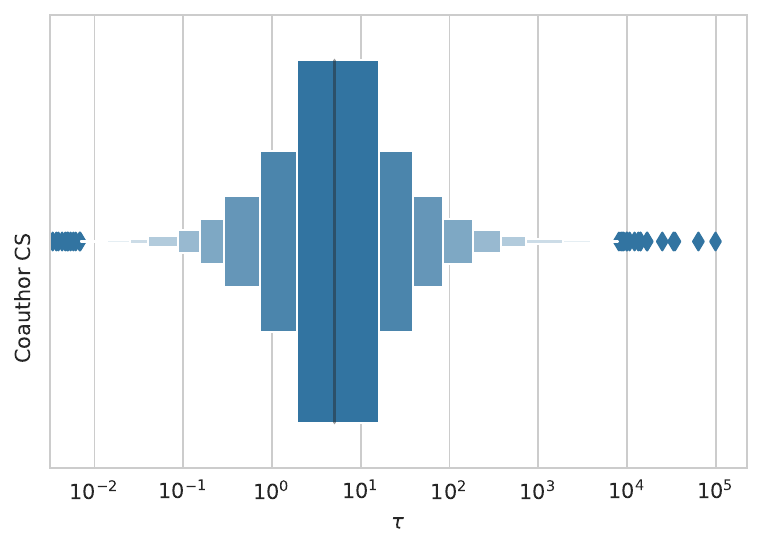}
         \caption{Coauthor CS}
     \end{subfigure}
     \hfill
     \begin{subfigure}[b]{0.32\textwidth}
         \centering
         \includegraphics[width=\textwidth]{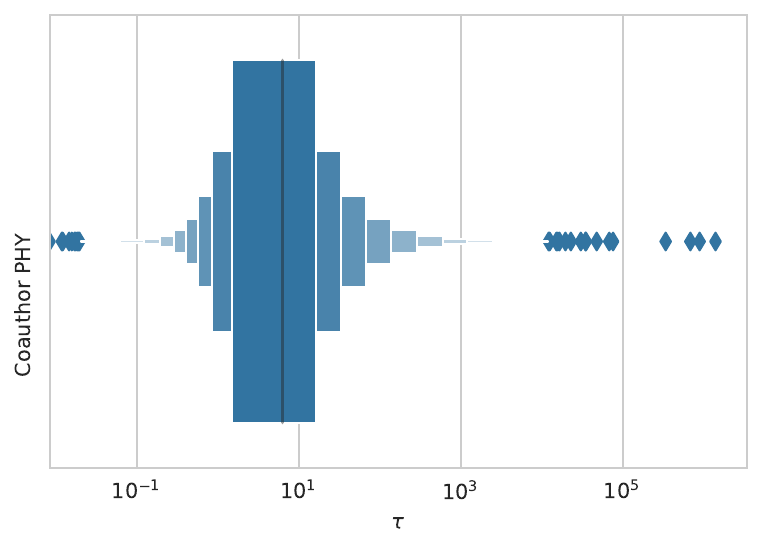}
         \caption{Coauthor PHY}
     \end{subfigure}
     \vfill
     \begin{subfigure}[b]{0.32\textwidth}
         \centering
         \includegraphics[width=\textwidth]{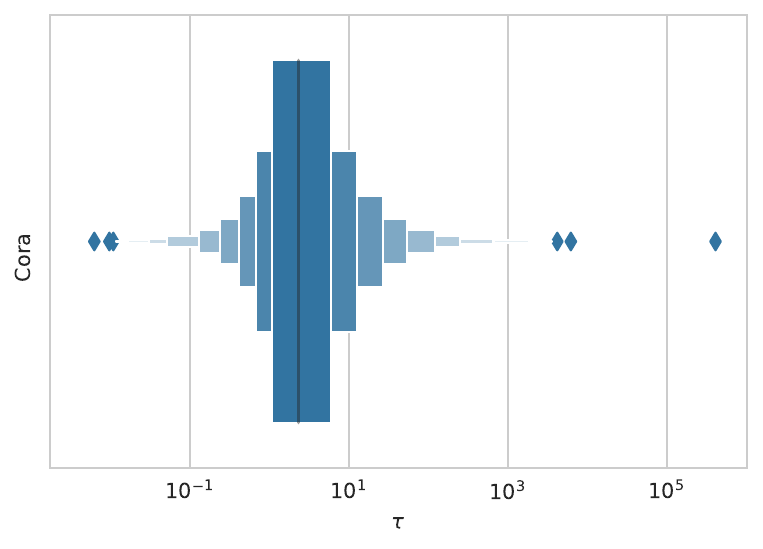}
         \caption{Cora}
     \end{subfigure}
     \hfill
     \begin{subfigure}[b]{0.32\textwidth}
         \centering
         \includegraphics[width=\textwidth]{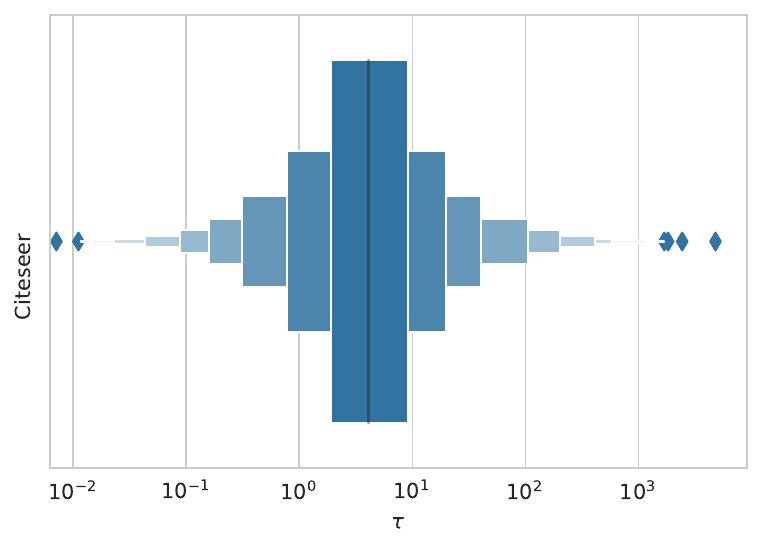}
         \caption{Citeseer}
     \end{subfigure}
     \hfill
     \begin{subfigure}[b]{0.32\textwidth}
         \centering
         \includegraphics[width=\textwidth]{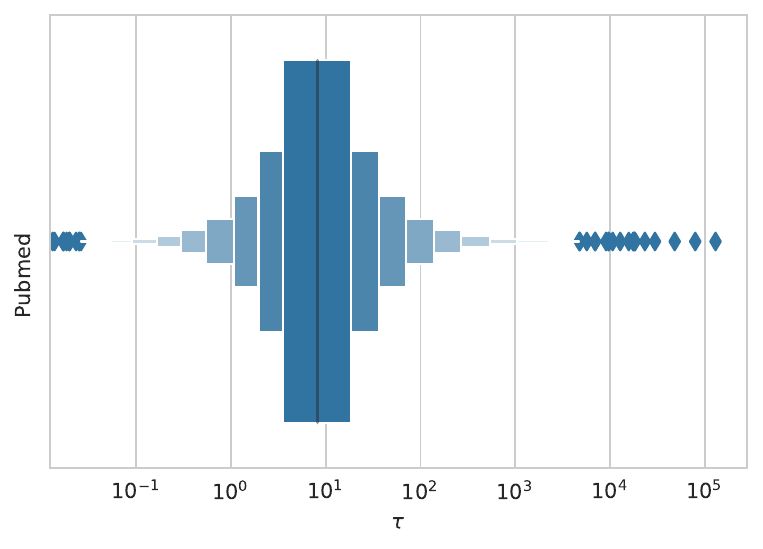}
         \caption{Coauthor PHY}
     \end{subfigure}
     \vfill
     \begin{subfigure}[b]{0.32\textwidth}
         \centering
         \includegraphics[width=\textwidth]{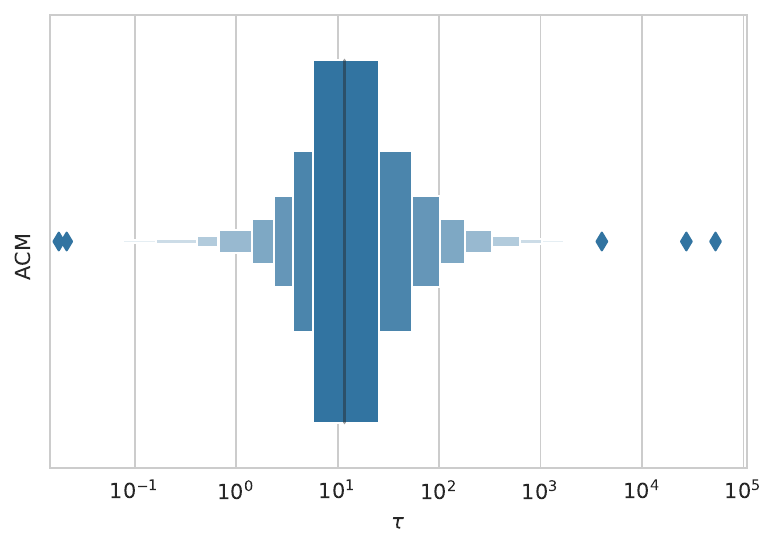}
         \caption{Cora}
     \end{subfigure}
     \hfill
     \begin{subfigure}[b]{0.32\textwidth}
         \centering
         \includegraphics[width=\textwidth]{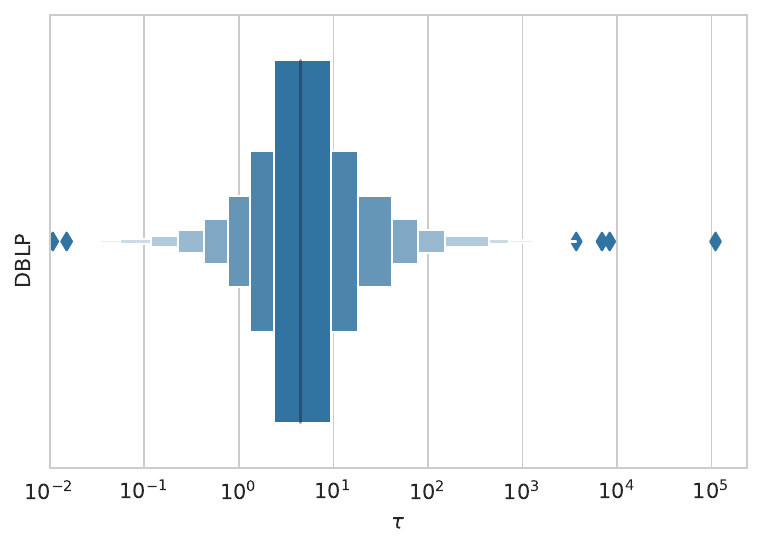}
         \caption{Citeseer}
     \end{subfigure}
    \caption{Empirical evaluation of $\tau^{0}$ distribution across 8 datasets.}
    \label{fig: tau dist quantile}
\end{figure*}

From Eq.8, we empirically evaluate $\mathbb{E}_{\substack{x^- \sim p_x^-} } h(x,x^-)$ by $\widehat{\mathbb{E}}_{x^{-} \in \mathcal{X}}h(x,x^-) $, i.e. the average of all negative samples of each sample $x$ across all samples $\mathcal{X}$. Similar empirical approximation $\widehat{\mathbb{E}}_{x^{+} \in \mathcal{X}}h(x,x^+) $ will be applied to $\mathbb{E}_{\substack{x^+ \sim p_x^+} } h(x,x^+)$.
An empirical evaluation of $\tau^{0}$ across 8 datasets is provided in Figure~\ref{fig: tau dist quantile}. The values of $\tau^{0}$ remain small values for most samples as in Table~\ref{tab: tau dist}.

\begin{small}
\begin{table}[tbh]
    % \centering
    \small
    \caption{DMAT-i empirical evaluation of $\tau^{0}$ distribution with several quantiles: Q1 (25\%) and Q3 (75\%)}
    \begin{adjustbox}{max width=\columnwidth,center}
    \begin{tabular}{lrrrrrrrr}
    \hline
    Quantile  & Coauthor CS & Coauthor PHY & Amazon Photo &	ACM & Citeseer & DBLP &	Cora & Pubmed \\ 
    \hline
    Q1  & $ 1.90 $  & $ 1.48 $  & $ 1.26 $   & $ 5.76 $ & $ 1.91 $ & $ 2.38 $ & $ 1.06 $ & $ 3.52 $ \\
    Median & $ 5.11 $  & $ 6.21 $  & $ 2.74 $ & $ 11.67 $ & $ 4.09 $ & $ 4.50 $ & $ 2.32 $ & $ 8.09 $ \\
    Q3  & $ 16.18 $  & $ 16.35 $  & $ 5.70 $ & $ 25.47 $ & $ 9.14 $ & $ 9.47 $ & $ 5.88 $ & $ 18.37 $ \\
    \hline
    \end{tabular}
    \end{adjustbox}
    \label{tab: tau dist}
\end{table}   
\end{small}

% \begin{table}[tbh]
%     % \centering
%     \caption{Empirical evaluation of $\tau^{0}$ Distribution}
%     \begin{adjustbox}{center}
%     \begin{tabular}{lrrrrrrrr}
%     \hline
%     Quantile  & Coauthor CS & Coauthor PHY & Amazon Photo &	ACM & Citeseer & DBLP &	Cora & Pubmed \\ 
%     \hline
%     Q1  & $ 1.90 $  & $ 1.48 $  & $ 1.26 $   & $ 5.76 $ & $ 1.91 $ & $ 2.38 $ & $ 1.06 $ & $ 3.52 $ \\
%     Median & $ 5.11 $  & $ 6.21 $  & $ 2.74 $ & $ 11.67 $ & $ 4.09 $ & $ 4.50 $ & $ 2.32 $ & $ 8.09 $ \\
%     Q3  & $ 16.18 $  & $ 16.35 $  & $ 5.70 $ & $ 25.47 $ & $ 9.14 $ & $ 9.47 $ & $ 5.88 $ & $ 18.37 $ \\
%     \hline
%     \end{tabular}
%     \end{adjustbox}
%     \label{tab:tau dist}
% \end{table}   

\subsection{Empirical Evaluation of representation Negative Hardness}

Our train objectives of DM(A)T and DMAT-i are hardness-ware loss functions with the strength of penalties on hard negative samples~\cite{wang_2021_CL_behavior}. In supervised machine learning, using hard (true negative) samples can accelerate a learning method by correcting mistakes faster. In representation learning, we consider a pair of negative examples as 
being informative if their latent representation are mapped nearby but should be actually be 
far apart.  

To investigate the hardnes-awareness of DMAT-i, we studied the distribution of similarity scores $Z_{i}^{T}Z_{j}$ as in Algorithm 1 among negative sample pairs. The 
idea is that  a negative sample pair with a higher score will be more difficult to discriminate and thus considered to be hard. Hard negative samples have been  proven to accelerate the 
learning process,  while too much hardness can also degrade the performance. 

\begin{figure*}
     \centering
     \begin{subfigure}[b]{0.32\textwidth}
         \centering
         \includegraphics[width=\textwidth]{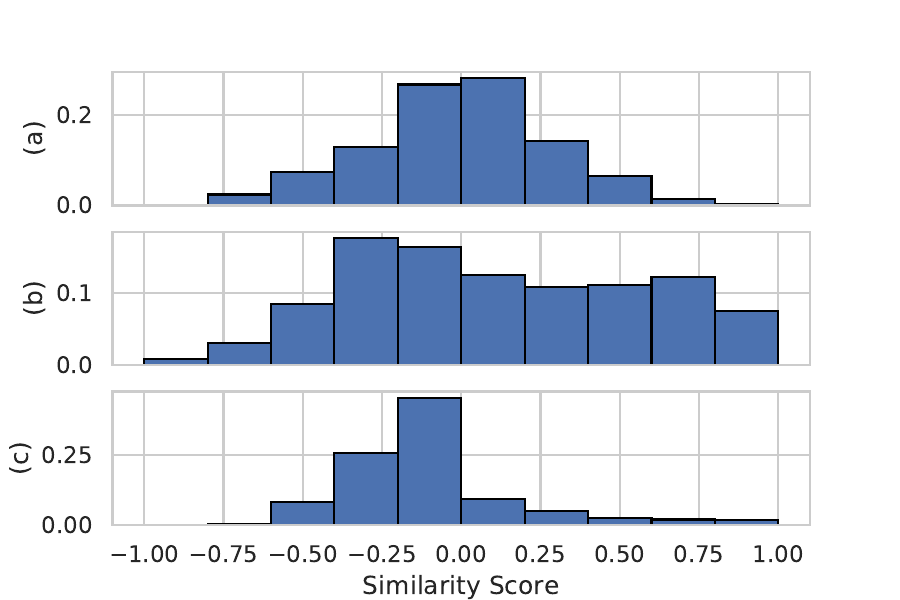}
         \caption{Amazon Photo}
     \end{subfigure}
     \hfill
     \begin{subfigure}[b]{0.32\textwidth}
         \centering
         \includegraphics[width=\textwidth]{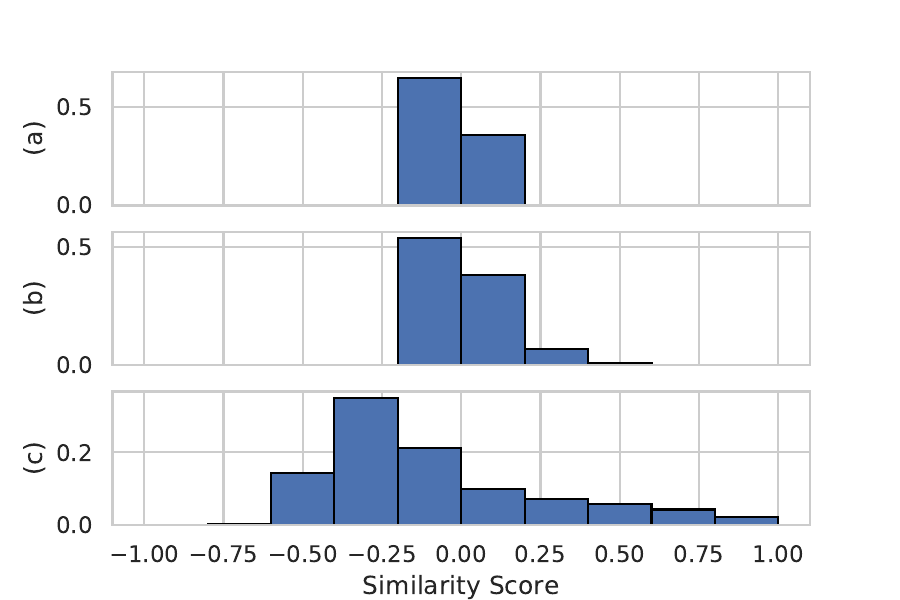}
         \caption{Coauthor CS}
     \end{subfigure}
     \hfill
     \begin{subfigure}[b]{0.32\textwidth}
         \centering
         \includegraphics[width=\textwidth]{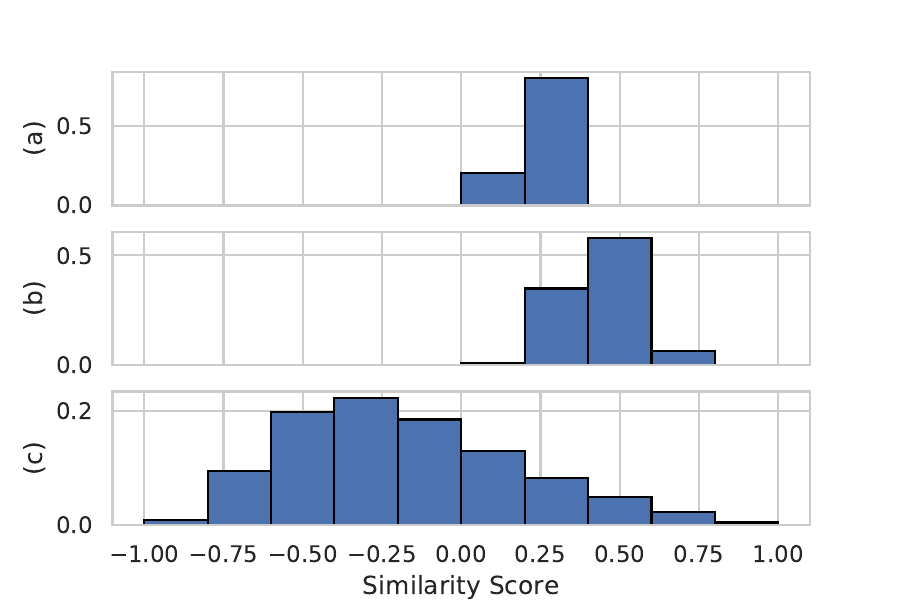}
         \caption{ACM}
     \end{subfigure}
     \vfill
     \begin{subfigure}[b]{0.32\textwidth}
         \centering
         \includegraphics[width=\textwidth]{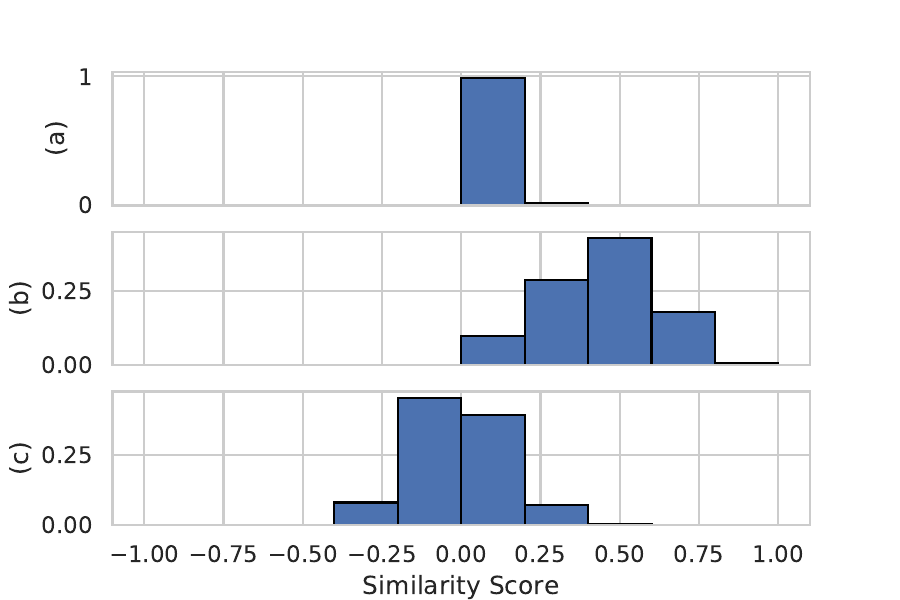}
         \caption{Cora}
     \end{subfigure}
     \hfill
     \begin{subfigure}[b]{0.32\textwidth}
         \centering
         \includegraphics[width=\textwidth]{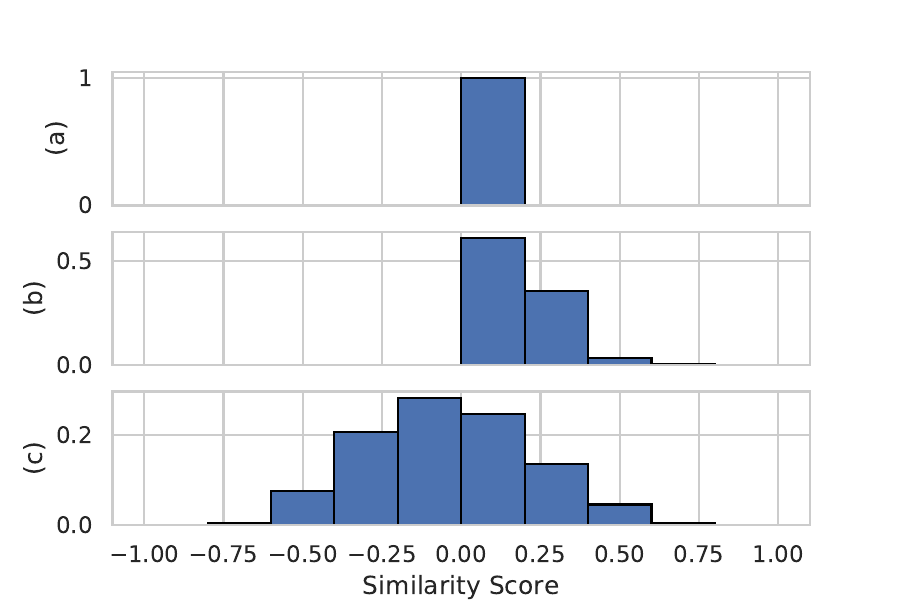}
         \caption{Citeseer}
     \end{subfigure}
     \hfill
     \begin{subfigure}[b]{0.32\textwidth}
         \centering
         \includegraphics[width=\textwidth]{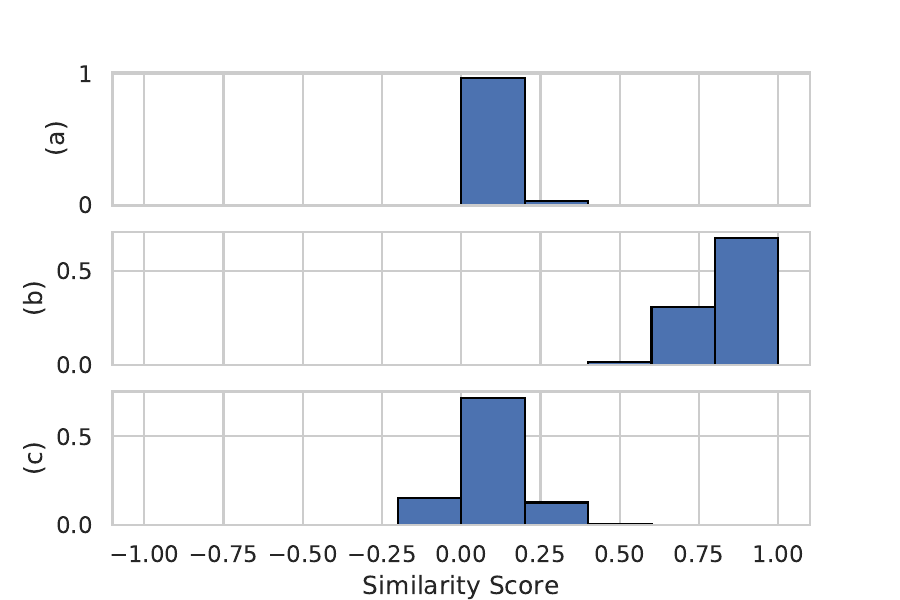}
         \caption{PubMed}
     \end{subfigure}
     \vfill
     \begin{subfigure}[b]{0.32\textwidth}
         \centering
         \includegraphics[width=\textwidth]{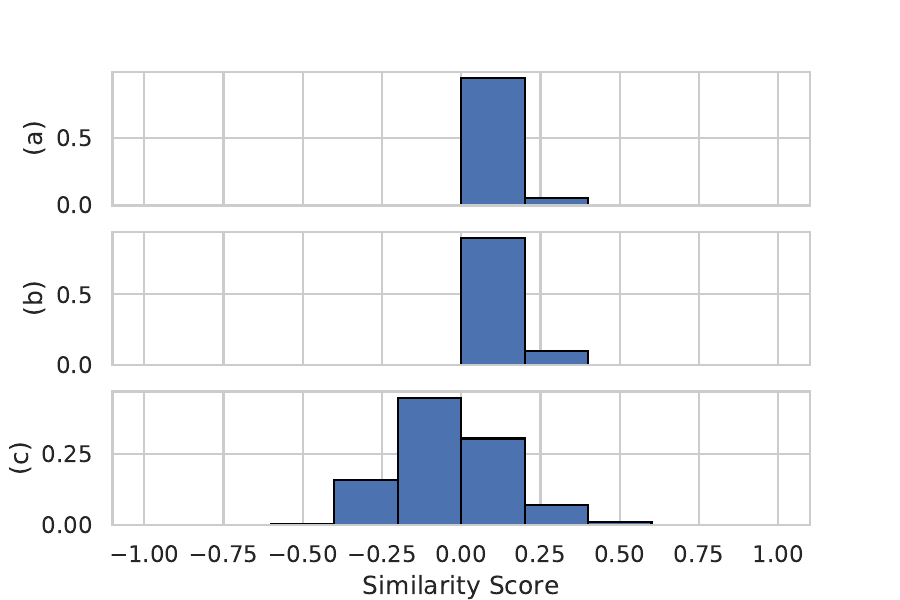}
         \caption{DBLP}
     \end{subfigure}
    \caption{Hard negative sample pair fraction distribution across 7 datasets. (a) raw node attributes; (b) smoothed features; (c) embedding after training}
    \label{fig: repre hardness}
\end{figure*}

We perform an empirical evaluation of negative hardness in 7 out of 8 datasets (except Amazon PHY whose large size makes such evaluation unfeasible).
As shown in \textbf{Figure} \ref{fig: repre hardness}, the smoothed features contain many 
more hard negative sample pairs than raw features. For most datasets,  the hardness concentrates
more  within a range  of similarity scores $[0.25, 0.50]$, therefore more hard negative samples with \textit{mild hardness} are included due to graph filtering. The embedding after training process shows a significant reduction of hardness, which indicates that our model is able to discriminate most of the negative pairs through continuous self-optimization.

\clearpage

\section{Generalized PageRank and Laplacian Smoothing Filters}
\label{subsec:scaling} 

We present how $\mathcal{X}$ is calculated for the purpose of  
our approach. As a background, 
Graph convolutional Network (GCN) has been a popular technique  for  incorporating graph structure information for representation learning. However, it has been shown to  suffer  from over-smoothing issues, leading to  degraded learning performance~\cite{Li_Deeper_into_GCN_2018}.
As a remedy for such issue, graph smoothing filters  have been successful, with specific instances being Adaptive Graph Convolution (AGC)~\cite{ijcai2019_AGC}, Simple Spectral Graph Convolution  (SSGC)~\cite{ssgc_2021} and Adaptive Graph Encoder (AGE)~\cite{Cui_AGE_2020}. 
We observe that Laplacian smoothing can be approximated by Generalized Page Rank (GPR), which can be computed by a scalable and parallelizable random-walk based algorithm \cite{cwdlydw2020gbp}.

Returning to the notation introduced at the beginning of the section, let $D$ be the diagonal degree matrix and
the graph Laplacian matrix is defined as $L=D-A$. Let $\widetilde{A} = A + I_{N}$ and $\widetilde{D}$ represent the augmented adjacency matrix with self-loops added and the corresponding diagonal degree matrix. 
With $\widetilde{L}=\widetilde{D}-\widetilde{A}$ as the Laplacian matrix corresponding to $\widetilde{A}$, the generalized Laplacian smoothing filter ${\cal{H}}$ \cite{Taubin_Laplacian_smoothing_1995} used by Cui {\em et al.}~\cite{Cui_AGE_2020} can be stated as: 

\vspace{-5pt}
\begin{equation}
\label{eq:Symmetric Laplacian smoothing matrix}
    \begin{split}
        \hat{Y} = {\mathcal{H}}\widetilde{X} = (I-\gamma\widetilde{D}^{-\frac{1}{2}}\widetilde{L}\widetilde{D}^{-\frac{1}{2}})\widetilde{X} \\
    \end{split}
\end{equation} 

% GCN turns out to be  a  limited  form of Laplacian smoothing by setting $\gamma=1$.
Now, stacking $L$ Laplacian smoothing filters from Eq.\ref{eq:Symmetric Laplacian smoothing matrix} yields:

\vspace{-5pt}
\begin{equation}
    \label{eq:AGE smoothing}
    \begin{split}
       \hat{Y}={\cal{H}}^{L}\widetilde{X}=(\gamma\widetilde{D}^{-\frac{1}{2}}\widetilde{A}\widetilde{D}^{-\frac{1}{2}}+(1-\gamma)I_{n})^{L}\cdot \widetilde{X}
    \end{split}
\end{equation}

While Laplacian smoothing can serve as a filter that addressed over-smoothing 
issues of GCN, GPR provides a mechanism to perform scalable computations. 
To this end,  Chen {\em et al.}~\cite{cwdlydw2020gbp} designed a localized bidirectional propagation algorithm as an unbiased estimate of the Generalized PageRank matrix ($P$)~\cite{PanLi_2019seed_expansion} as below: 

\vspace{-5pt}
\begin{equation}
    \label{eq:pagerank}
    \begin{split}
       P = \sum\limits_{l=0}^{L}w_{l}T^{l} = \sum\limits_{l=0}^{L}w_{l}(\widetilde{D}^{r-1}\widetilde{A}\widetilde{D}^{r})^{l}\cdot \widetilde{X}
    \end{split}
\end{equation}

with $r \in [0, 1]$ as the convolution coefficient, $w_{l}$'s as the weights of different order convolution matrices satisfying $\sum^{\infty}_{l=0}w_{l} \leq 1$ and $T^{l}=(\widetilde{D}^{r-1}\widetilde{A}\widetilde{D}^{r})^l \cdot X$ as the $l-th$ step propagation matrix.

Eq. \ref{eq:AGE smoothing} can be generalized into GPR in Eq.\ref{eq:pagerank} by setting $r=0.5$ and manipulating the weights $w_{l}$ to simulate various diffusion processes. 
Therefore, we use the parallelizable bidirection propagation algorithm from Chen {\em et al.}~\cite{cwdlydw2020gbp} as a highly scalable graph smoothing filter to ease the downstream  tasks. By setting $w_l = \alpha(1-\alpha)^{l}$ with $\alpha$ as the teleport probability \cite{Klicpera2019_APPNP}, we focused on one setup  where $\bf{P}$ becomes the PPR (Personalized PageRank) as used in APPNP ~\cite{Klicpera2019_APPNP}. The hyper-parameter $\alpha$ will adjust the size of the neighborhood for node information aggregation~\cite{Klicpera2019_APPNP} such that we can balance the needs of preserving locality and leveraging information of nodes from further distances. PPR can avoid over-smoothing issues even when aggregating attributes from infinite layers of neighboring nodes~\cite{Klicpera2019_APPNP}.

\section{Complexity Analysis}

\begin{table*}
    \centering
    \caption{Time and Space complexity during Training on the GPU O($\cdot$), with $N$ as the number of nodes, $E$ as the number of edges, $d$ as the feature dimension, $L$ as the number of neural network layers, $\cal{B}$ as the batch size.}
    \begin{adjustbox}{max width=\linewidth}
    \begin{tabular}{l|r|r|r|r|r|r|r|r}
    \hline
    Complexity & GCA~\cite{zhu2021_GCA} & DGI~\cite{velickovic_2019_DGI} & SDCN \cite{Bo_2020_structural_deep} & ProGCL \cite{xia2022_progcl} & AGC \cite{ijcai2019_AGC} & SSGC \cite{ssgc_2021} & AGE~\cite{Cui_AGE_2020} & DM(A)T \\
     \hline
    Computation Cost & $L\cdot E \cdot d + L\cdot N \cdot d^{2}$ & $L\cdot E \cdot d + L\cdot N \cdot d^{2}$ & $L\cdot E \cdot d + L\cdot N \cdot d^{2}$ & $L\cdot E \cdot d + L\cdot N \cdot d^{2}$ & $E \cdot d + N \cdot d$ & $E\cdot d+N \cdot d$ & $E\cdot d+N \cdot d$ & $L\cdot {\cal{B}} \cdot d^{2}$ \\
    Memory Cost & $N\cdot d+ E +L\cdot d^{2}$ & $N\cdot d+ E +L\cdot d^{2}$ & $N\cdot d+ E +L\cdot d^{2}$ & $N\cdot d+ E +L\cdot d^{2}$ & $N\cdot d+ E $ & $N\cdot d+ E $ & $N\cdot d+ E $ & ${\cal{B}}\cdot d+L\cdot d^{2}$ \\
     \hline
    \end{tabular} 
    \end{adjustbox}
    \label{tab: Complexity}
\end{table*}

For this analysis, let $N$ be  the number of nodes, $L$ be  the number of neural network layers, $E$ be the total number of edges, $d$ be the number of features and $\cal{B}$ be the batch size. For simplicity, we assume number of features for all layers is fixed as $d$. In Table~\ref{tab: Complexity}, 
for  DM(A)T, the  computational cost is $O(L\cdot {\mathcal{B}}\cdot d^{2}+N\cdot d)$. Here,  $O(L\cdot {\mathcal{B}}\cdot d^{2})$ corresponds  to $L$ layers of matrix multiplication for DNN module during the training.
The memory cost is $O({\cal{B}}\cdot d+L\cdot d^{2})$ 
-- ${\cal{B}}\cdot d$ relates to saving a batch of input $\mathcal{X}$ and $L\cdot d^{2}$ is for storing model learning parameters $\{W^{(l)}\}_{l=1}^{L}$. 

In contrast, all the other baseline clustering frameworks include an expensive sparse matrix-matrix multiplication $\widetilde{A}\cdot \widetilde{X}$, where $\widetilde{A}$ refers to the adjacency matrix of the 
entire graph and $\widetilde{X}$ is the feature matrix. GCA, DGI, SDCN and ProGCL frameworks use a traditional GCN module,  so, they have an identical complexity, 
which turns out to be most expensive. 
AGC, SSGC, AGE are similar -- 
 the  computational cost is $O(E\cdot d+N\cdot d)$ since each $\widetilde{T}\cdot \widetilde{X}$ costs $E\cdot d$  and 
$N\cdot d$ is the cost of summation over filters and adding features. For all these three frameworks, the memory cost is $O(N\cdot d + E)$.  
Overall, since $E \gg N \gg d$ and the batch size ${\cal{B}}$ can be defined by users based on resource 
availability, the training process of our framework can be easily fit onto GPU memory for large-scale graphs. 

\section{Related Work}

\noindent\textbf{Attributed Graph Representation Learning.}
Graph Autoencoder (GAE) based models \cite{kipf2016variational_VGAE,Pan_2018_AVGAE,wang_mgae_2017}  learn 
node embeddings that can recover either node features or the adjacency matrix. 
%through their  architectures. 
Adaptive Graph Convolution (AGC) \cite{ijcai2019_AGC}   uses high-order graph convolution to capture global cluster structure and adaptively selects   the appropriate order for different graphs. Simple Spectral Graph Convolution (SSGC)  \cite{ssgc_2021}  is a variant of GCN that  exploits a modified Markov Diffusion Kernel.  Adaptive Graph Encoder (AGE) \cite{Cui_AGE_2020} combines a customized Laplacian smoothing filter with an adaptive encoder  to strengthen the filtered features for better node embeddings. In using GCN modules for node clustering,
SDCN~\cite{Bo_2020_structural_deep} is a 
self-supervised method. 
\cite{zhao2021_graph} proposed a graph debiased contrastive learning approach to jointly perform representation learning and clustering. GCA~\cite{zhu2021_GCA} is a graph contrastive
representation learning method with adaptive augmentation. ProGCL~\cite{xia2022_progcl} boosts the graph contrastive learning by estimating the probability of a negative being true. DCRN~\cite{liu2022_deep} improves the representation discriminative capability for node clustering by reducing information correlation.

\noindent\textbf{Deep Metric Learning.} 
%As mentioned previously, here one trains the networks to learn  feature embeddings where similar examples are mapped close to each other and dissimilar examples are mapped farther apart. 
\cite{bromley_1993_signature} laid the foundation of this area motivated by 
signature verification. \cite{chopra_2005_learning} discriminatively trains the network for face verification via contrastive loss, \cite{Schroff_2015_FaceNetAU} learns a unified embedding for face clustering and recognition through triplet loss, and 
\cite{cui_2016_fine} focuses on  visual categorization. \cite{Sohn_N-pair_loss_objective_2016} proposed (N+1)-tuplet loss for a variety of visual recognition tasks. \cite{yoshida_2021_distance} proposes a supervised distance metric learning for graph classification. 
It should be noted that  the contrastive loss here aims to ensure that the distance between examples from different classes is larger than a certain margin~\cite{Sohn_N-pair_loss_objective_2016}. 
The methods we next discuss under constrastive learning have a different emphasis. 
%which is different from the concept as defined in contrastive learning as introduced in the following. 

\noindent\textbf{Contrastive Learning.}
As one major branch of self-supervised learning, contrastive learning, 
when combined with augmentation techniques,  has achieved state-of-the-art performance in visual representation learning tasks~\cite{oord_2019_representation,hjelm_2019_DIM,Tian_multi_coding_2020,chen_simple_framework_contastive_2020,contrastiveLoss_wu2018}. This idea has also been applied for node-level representation learning for graph-structured data~\cite{velickovic_2019_DGI,zhu2021_GCA}. Deep Graph Infomax (DGI) \cite{velickovic_2019_DGI} achieves advanced node classification performance by extending  a contrastive learning mechanism (from precursor work Deep InfoMax~\cite{hjelm_2019_DIM}). 
GCA~\cite{zhu2021_GCA} applied the contrastive loss from~\cite{chen_simple_framework_contastive_2020} to maximize agreements at the node level and performed well on node classification tasks.  
We later discuss how our approach differs and also perform  extensive experimental comparison.

\section{Details for Experimental Setup}

Our DM(A)T framework is implemented in PyTorch 1.7 on CUDA 10.1, whereas   the graph filtering procedure is in C++. Our experiments are performed on nodes with a dual Intel Xeon 6148s @2.4GHz CPU and dual NVIDIA Volta V100 w/ 16GB memory GPU and 384 GB DDR4 memory. 
Graph filtering is executed on CPU while tuplet loss based training process is performed on a single GPU. We applied a AdamW optimization method with a decoupled weight decay regularization technique \cite{loshchilov2019decoupled_weight_decay}. 

\subsection{Datasets Details}

\begin{footnotesize} 
\begin{table}[tbh]
    % \centering
    \caption{Datasets Statistics}
    \begin{adjustbox}{max width=\columnwidth,center}
    \begin{tabular}{lrrrr}
    \hline
    dataset  & Nodes & Classes & Features & Edges \\ 
    \hline
    ACM       & $ 3025  $  & $ 3 $  & $ 1870 $   & $13,128$ \\
    DBLP      & $ 4058 $   & $ 4 $  & $ 334 $    & $3528$ \\
    Citeseer  & $ 3327  $  & $ 6 $  & $ 3703 $   & $4732$ \\
    Cora      & $ 2708  $  & $ 7 $  & $ 1433 $   & $5429$ \\
    Pubmed    & $ 19717  $  & $ 3 $  & $ 500 $   & $44,338$ \\
    Amazon Photo  & $ 7650 $   & $ 8 $ & $ 745 $    &  $71,831$ \\
    Coauthor CS     & $ 18333 $  & $ 15 $ & $ 6805 $  &  $81,894$   \\
    Coauthor PHY     & $ 34493 $  & $ 5 $ & $ 8415 $  &  $247,962$  \\
    % Ogb-arxiv  & $ 169343 $  & $ 40 $ & $ 128 $  &  $ 1,166,243	 $  \\
    % Reddit   & $ 232965 $  & $ 41 $ & $ 602 $  &  $114,615,892$  \\
    % Friendster   &  $6.5 \times  10^7$   & $ 7 $ & $ 40 $  &  $1.8 \times 10^9$  \\
    \hline
    \end{tabular}
    \end{adjustbox}
    \label{tab:Datasets info}
\end{table}   
\end{footnotesize}  

As in Table~\ref{tab:Datasets info} \textbf{ACM} \cite{Bo_2020_structural_deep}  is a paper network from ACM  --  nodes correspond 
to  paper, edges represent common author,   features are bag-of-words of keywords, 
and the class labels are the research areas.  \textbf{DBLP} \cite{Bo_2020_structural_deep}  is an author network --  
nodes represent authors, edges represent co-authorship,   features are the bag-of-words  of keywords, 
and class labels are the author's research fields based on the conferences they submitted papers to. \textbf{Amazon Photo}~\cite{image_recommend_McAuley_2015} is a segment of Amazon co-purchase graph -- nodes are goods, edges indicate two goods are frequently bought together, features are bag-of-words from product reviews and class labels represent product categories. We also include three 
Citation Networks\cite{Sen_2008_citation_networks}, i.e.,  {\bf  Cora}, {\bf Pubmed} and {\bf Citeseer}  here,  the nodes correspond to  papers, edges are citation links,  features are bag-of-word of  abstracts, 
and labels are paper topics. For consistent comparison, we use these datasets without row-normalization as described in \cite{kipf2017semi_GCN}.
We include two  Microsoft Co-authorship  datasets \cite{shchur2019pitfalls}, which are  {\bf Coauthor CS}  and {\bf Coauthor PHY},  based on Microsoft Academic Graph for computer science and physics, respectively -- here,  nodes represent authors, edges represent co-authorship, 
features are paper keywords, and  class labels indicate the  most common fields of study.  

\subsection{Scalability of Representation Construction settings}  
Synthetic datasets were used to evaluate scalability of DMAT-i and other  frameworks.
We used PaRMAT \cite{PaPMAT_2015} to generate undirected synthetic graphs of growing size with edge count set as 20 times the number of nodes and a random feature matrix with a dimension of 1000. For each clustering method, we performed 5-epoch training, repeated  each experiment 5 times, and report average times. 

\subsection{DMAT Hyper-parameter Settings}\label{RwSL hyper-parameter}

\begin{table*}
    \centering
    \caption{ DMAT hyper-parameter settings on 8 datasets}
    \begin{adjustbox}{width=\textwidth}
    \begin{tabular}{lrrrrrrrr}
    \hline
    Hyper-parameters  & ACM     & DBLP      & Cora      & CiteSeer  & PubMed  & Amazon Photo & Coauthor CS & Coauthor PHY  \\
    \hline
    learning\_rate  & $ 1e-3 $  & $ 1e-3 $  & $ 1e-4 $  & $ 1e-4 $  & $ 1e-5 $  & $ 8e-5 $   & $ 1e-5 $  & $ 2e-5 $  \\
    architecture    & 256-128   & 256-256   & 256-128   & 256-512   & 256-256   & 512-512    & 256-512   & 256-512   \\
    $\tau$          & $ 2.0 $   & $ 2.0 $   & $ 1.0 $   & $ 4.0 $   & $ 0.8 $   & $ 2.0 $    & $ 1.2 $   & $ 0.5 $   \\
    n\_epochs       & $ 400 $   & $  300 $  & $ 300 $   & $ 400 $   & $ 200 $   & $ 500 $    & $ 400 $   & $ 400 $   \\
    mask\_fraction   & $ 0.6 $   & $ 0.2 $   & $ 0.08 $  & $ 0.2 $   & $ 0.2 $   & $ 0.1 $    & $ 0.4 $   & $ 0.1 $   \\
    view\_num       & $ 4 $     & $ 4 $     & $ 3 $     & $ 4 $     & $ 2 $     & $ 4 $      & $ 5 $     & $ 5 $     \\
    weight\_decay   & $ 0.02 $  & $ 0.05 $  & $ 0.02 $  & $ 0.05 $  & $ 0.05 $  & $0.1$      & $ 0.05 $  & $ 0.05 $  \\
    batch\_size     & $ 512 $   & $ 512 $   & $ 512 $   & $ 512 $   & $ 512 $   & $ 256 $    & $ 512 $   & $ 512 $   \\
    $\alpha$        & $ 0.4 $   & $ 0.6 $   & $ 0.1 $   & $ 0.4 $   & $ 0.01 $  & $ 0.03 $   & $ 0.1 $   & $ 0.08 $  \\
    $r_{max}$       & $ 1e-5 $  & $ 1e-4 $  & $ 1e-6 $  & $ 1e-5 $  & $ 1e-5 $  & $ 1e-6 $   & $ 1e-5 $  & $ 1e-5 $  \\
    rrz             & $ 0.4 $   & $ 0.5 $   & $ 0.4 $   & $ 0.4 $   & $ 0.4 $   & $ 0.5 $    & $ 0.4 $   & $ 0.4 $   \\
    \hline
    \end{tabular}
    \end{adjustbox}
    \label{tab: DMAT datasets hyper-parameters}
\end{table*}

Detailed hyper-parameters settings are included in Table \ref{tab: DMAT datasets hyper-parameters}. Learning\_rate corresponds to the learning rates during DNN based encoder training with a contrastive loss. During the augmentation of node embeddings to produce multiple views, mask\_fraction is the portion of columns to mask and view\_num is the number of generated views. $\tau$ is the temperature parameter of contrastive loss \cite{zhu2021_GCA}. And n\_epochs is the number of iterations for DNN training. The architecture describes the number of neurons on each layer of the encoder. We applied a decoupled weight decay regularization \cite{loshchilov2019decoupled_weight_decay} resulting in the factor weight\_decay. The size of batch during mini-batch training is controled by batch\_size. The last three parameters are from GnnBP framework \cite{cwdlydw2020gbp}, $\alpha \in  (0, 1)$ is teleport probability defined in
Personalized PageRank weights $(w_{l}=\alpha(1-\alpha)^{l})$; $r_{max}$ is the threshold during reverse push propagation from the feature vectors; $rrz$ is the convolutional coefficient.

\subsection{Baseline Justification and Source Codes}\label{baseline}

In justifying our choice of baselines for the node clustering task, we observe that methods that  utilize both node features and graph structure  achieve a significant improvement over other approaches that only exploit one of them. Earlier attributed graph embedding frameworks such as \textbf{GAE} and \textbf{VGAE}\cite{kipf2016variational_VGAE}, \textbf{ARGE}  and  \textbf{ARVGE}\cite{Pan_2018_AVGAE} were outperformed by either or both of \textbf{AGC} and \textbf{SSGC}. In addition, \textbf{SSGC} outperformed \textbf{SGC}~\cite{SGC_2019}. Deep Embedded Clustering (\textbf{DEC}) \cite{Xie_DEC_2016}, Improved Deep Embedded Clustering (\textbf{IDEC})\cite{ijcai_IDEC_2017}, \textbf{MGAE}\cite{wang_mgae_2017}, and Deep Attentional Embedded Graph Clustering (\textbf{DAEGC})\cite{wang_DAEGC_2019} were outperformed by \textbf{SDCN} as baselines, \textbf{GALA}~\cite{GALA_park_2019} was outperformed by \textbf{AGE}~\cite{Cui_AGE_2020} and finally \textbf{MVGRL}~\cite{hassani2020_mvgrl} was outperformed by \textbf{GCA}~\cite{zhu2021_GCA}.

All baseline codes used are summarized  in Table~\ref{tab:baseline codes}.

\begin{table}[tbh]
\begin{small} 
    \centering
    \caption{URLs of GnnBP precomputation and baseline codes}
    \label{tab:baseline codes}
   \begin{adjustbox}{width=0.75\columnwidth,center}
    \begin{tabular}{cl}
    \hline
    Framework  & URL \\
    \hline
    GnnBP    & \url{https://github.com/chennnM/GBP} \\
    DeepWalk    & \url{https://github.com/phanein/deepwalk} \\
    SDCN    & \url{https://github.com/bdy9527/SDCN} \\
    ProGCL    & \url{https://github.com/junxia97/ProGCL} \\ 
    AGC    & \url{https://github.com/karenlatong/AGC-master} \\
    SSGC    & \url{https://github.com/allenhaozhu/SSGC} \\
    AGE    & \url{https://github.com/thunlp/AGE} \\
    DGI    & \url{https://github.com/PetarV-/DGI} \\
    GCA    & \url{https://https://github.com/CRIPAC-DIG/GCA} \\
    \hline
    \end{tabular}
   \end{adjustbox}
    \end{small} 
\end{table}

\newpage
\clearpage

\bibliographystyle{splncs04}
\bibliography{ref}





% LNCS Homepage, \url{http://www.springer.com/lncs}. Last accessed 4

% \end{thebibliography}

% \input{input/appendix}

\end{document}